\documentclass[10pt,journal,compsoc]{IEEEtran}

\usepackage{stmaryrd}
\usepackage{adjustbox}
\usepackage{array}
\usepackage{multirow}
\usepackage{amsmath,bm}
\usepackage{amssymb}
\usepackage{pifont}
\usepackage{xspace}
\usepackage{tikz}
\usepackage{booktabs}
\usepackage{color}
\usepackage{graphicx}
\usepackage{algorithm, algpseudocode}
\usepackage{multicol}

\newcommand*{\eg}{\emph{e.g.}\@\xspace}
\newcommand*{\ie}{\emph{i.e.}\@\xspace}
\newcommand*{\etal}{\emph{et al.}\@\xspace}

\newcommand{\cmark}{\ding{51}}%

\newcommand\Tstrut{\rule{0pt}{2.2ex}}         

\makeatother
\usepackage[pagebackref=true,breaklinks=true,letterpaper=true,colorlinks,bookmarks=false]{hyperref}

\ifCLASSOPTIONcompsoc
  \usepackage[nocompress]{cite}
\else
  \usepackage{cite}
\fi

\ifCLASSINFOpdf
\else
\fi

\begin{document}
\title{Stylized Adversarial Defense}
\author{Muzammal~Naseer, 
        Salman~Khan, 
        Munawar~Hayat,
        Fahad~Shahbaz~Khan
        and~Fatih~Porikli
\IEEEcompsocitemizethanks{\IEEEcompsocthanksitem M. Naseer, S. Khan, and F. Khan are with Mohamed bin Zayed University of Artificial Intelligence, UAE\protect\\
E-mail: muz.pak@gmail.com
\IEEEcompsocthanksitem M. Naseer, and S. Khan are also with the Australian National University, Canberra, Australia.
\IEEEcompsocthanksitem F. Khan is also with Linköping University,
Sweden.
\IEEEcompsocthanksitem M. Hayat is with Monash University, Australia.
\IEEEcompsocthanksitem F. Porikli is with Qualcomm, USA.
}
}


\IEEEtitleabstractindextext{%
\begin{abstract}
Deep Convolution Neural Networks (CNNs) can easily be fooled by subtle, imperceptible changes to the input images. To address this vulnerability, adversarial training creates perturbation patterns and includes them in the training set to robustify the model. In contrast to existing adversarial training methods that only use class-boundary information (e.g., using a cross-entropy loss), we propose to exploit additional information from the feature space to craft stronger adversaries that are in turn used to learn a robust model.  Specifically, we use the \emph{style} and \emph{content} information of the target sample from another class, alongside its class-boundary information to create adversarial perturbations. We apply our proposed \emph{multi-task} objective in a deeply supervised manner, extracting multi-scale feature knowledge to create maximally separating adversaries. Subsequently, we propose a max-margin adversarial training approach that minimizes the distance between source image and its adversary and maximizes the distance between the adversary and the target image. Our adversarial training approach demonstrates strong robustness compared to state-of-the-art defenses, generalizes well to naturally occurring corruptions and data distributional shifts, and retains the model’s accuracy on clean examples. 
\end{abstract}

}

\maketitle

\IEEEdisplaynontitleabstractindextext

\IEEEpeerreviewmaketitle

\section{Introduction}
Although deep networks excel on a variety of learning tasks, they remain vulnerable to adversarial perturbations. These perturbations are imperceptible to humans, but significantly degrade the prediction accuracy of a trained model. Adversarial training \cite{madry2018towards} has emerged as a simple and successful mechanism to achieve robustness against adversarial perturbations. In this process,  blind-spots of the model are first found by crafting malicious perturbations and subsequently included in the training set to learn a robust model.

\begin{figure*}[t]
\centering
  	\centering
    \includegraphics[width=0.95\linewidth]{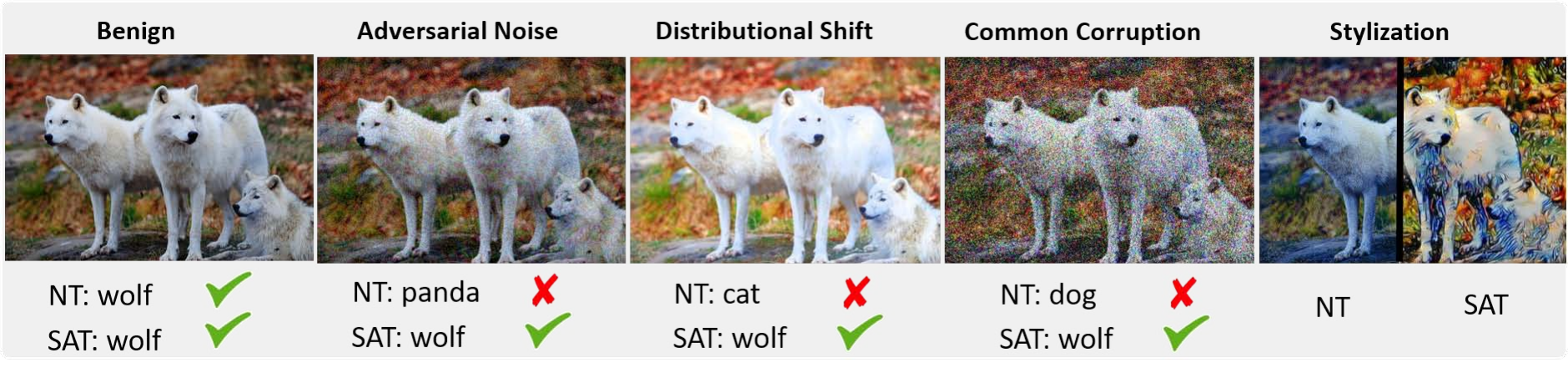}\
    \vspace{-1em}
  \caption{ A robust model trained with our proposed Stylized Adversarial Training (SAT) framework  generalizes  not only to adversarial noise but also handles naturally occurring distributional shifts (\eg, contrast change in the above example), common corruptions (e.g., sensor noise) and performs better stylization compared to a Naturally Trained (NT) model.
  }
 \label{fig:concept}
\end{figure*}

In this paper, we interpret adversarial training from a margin maximization perspective. We consider margin as the shortest distance from a data-point to the classifier’s boundary in the decision and perceptual (feature) spaces. Intuitively, the highest robustness can be achieved by learning a margin maximizing model that first crafts a maximally separated adversarial example and then readjusts the boundary to correctly classify such perturbed samples. However, in practice, this task turns out to be a nested max-min optimization problem, whose solution is non-trivial \cite{ding2018max}. Therefore, we propose an alternate way to maximize classifier margins. Our approach is motivated by the fact that adversarial training maximizes a lower bound on the classifier's margin. Towards this goal, our main idea is to identify a target image from a different class for guidance, and create perturbations that can push the source image towards the target in both feature and output spaces using a multi-task objective function.

In the pursuit of creating highly deceptive adversaries, we propose an attack based on multi-modal information including classifier’s boundary information, image style and visual content. In this manner, the perturbations cause significant changes to the intermediate feature as well as output decision space using a diverse multi-task attack objective based on multiple supervisory cues. The existing targeted attacks in the literature create adversaries by moving towards the least likely (or the most confusing) class \cite{wang2019bilateral, zhang2019joint, zhang2020adversarial}. In this manner, only the class boundary information is used to craft adversarial perturbations. Different to those, we create targeted adversaries by pushing the sample towards a randomly picked sample from a different class such that its style and content representations are also reshaped besides the output prediction. This is done by incorporating multi-scale information from the feature hierarchy in a deeply supervised manner \cite{lee2014deeplysupervised}.

Based on the proposed carefully crafted perturbations, we develop our Stylized Adversarial Training (SAT) approach to achieve robustness. Specifically, we enforce a margin-maximizing objective during adversarial training, which minimizes the distance between clean and perturbed images while maximizing the distance between clean image and target sample (used to create adversaries). The model thus learns corrective measures with respect to a reference sample from a different class, thereby enhancing the model's robustness. Since our attack objective uses supervision from the style and content of a target image from different class, it forces the perturbations to lie close to the natural image manifold. As a result, our adversarially trained model performs significantly better than other adversarial training approaches \cite{madry2018towards,Zhang2019theoretically,feature_scatter} on the clean images. Simultaneously, our proposed defense shows strong robustness against naturally occurring image degradations such as contrast changes, blurring and rain, that cause distributional shifts. We further demonstrate that the model trained with our proposed scheme performs much better on the style transfer task despite having less parametric complexity (see Fig.~\ref{fig:concept}).

The major contributions of our work are:
\begin{itemize}
\item We propose to set-up priors in the form of fooling target samples during adversarial training and propose a multi-task objective for adversary creation that seeks to fool the model in terms of image style, visual content as well as the decision boundary for the true class. 
\item Based on a high-strength perturbation, we develop a margin-maximizing (\emph{contrastive}) adversarial training procedure that maps perturbed image close to clean one and maximally separates it from the target image used to craft the adversary. 
\item With extensive evaluation, we demonstrate that transferring information from multi-task objectives helps us perform favorably well against the strongest adversarial training methods such as PGD based Adversarial Training (PAT) \cite{madry2018towards}, Trades \cite{Zhang2019theoretically} and Feature scattering \cite{feature_scatter}. 
\item Compared to conventional adversarial training, our approach does not cause a drop in clean accuracy, and performs well against the real-world common image corruptions \cite{hendrycks2019robustness}. We further demonstrate robustness and generalization capabilities of the proposed training regime when the underlying data distribution shifts (Sec.~\ref{subsec:results_robust_training_framework}).
\end{itemize}

\section{Related Works}

\textbf{Adversarial Training:} 
Training a model on adversarial examples can regularize it and increase its adversarial robustness. Goodfellow \etal \cite{goodfellow2014explaining} proposed a computationally fast adversary generation algorithm known as `Fast Gradient Sign Method' (FGSM). FGSM suffers from label leakage \cite{kurakin2016adversarial} that allows the model to overfit on FGSM's generated adversaries, hence hampering its adversarial generalization. Tamer \etal \cite{tramer2018ensemble} proposed to mitigate this issue by taking a small random step before running FGSM. Their attack is known as `Random Fast Gradient Sign Method' (RFGSM). 
Their method performs relatively better, but still suffers under iterative attacks \cite{carlini2017towards,madry2018towards}. Madry \etal \cite{madry2018towards} solved the overfitting problem by adversarially training models on iterative attack known as `Projected Gradient Descent' (PGD). 
PGD is an untargetted, label-dependent attack and models trained on PGD adversaries show significant robustness to the strongest white-box attacks \cite{carlini2017towards,kurakin2016adversarial}. 
However, PGD gains robustness at the cost of a significant drop in clean accuracy and lacks a clear mechanism to control the accuracy-robustness trade-off. This is where Zhang \etal \cite{Zhang2019theoretically} contributed and proposed a method to control the trade-off with an untargetted, label-independent (unsupervised) attack to create adversaries along with a surrogate clustering loss to minimize the model's empirical risk.

However, \cite{madry2018towards,Zhang2019theoretically} deploy iterative attacks which are computationally expensive and less scalable to high-dimensional datasets. Further, adversarial training done using untargetted attacks whether computationally expensive \cite{madry2018towards,Zhang2019theoretically} or efficient \cite{shafahi2019adversarial,Wong2020Fast} results in only a limited robustness. To improve it, \cite{feature_scatter} proposed a faster attack that operates in logit space in an unsupervised way to maximize optimal transport distance. Combined with label smoothing, their method produced state-of-the art results on SVHN, CIFAR10 and CIFAR100. In this work, we propose a conceptually simple and efficient adversarial training process, exploiting a multi-task loss that helps us perform favorably well against previous state-of-the-art methods.

\noindent \textbf{Augmentation based Adversarial Training:} Since traditional adversarial training results in a significant drop in clean accuracy, augmentation based methods have been proposed to overcome this limitation. \cite{lamb2019interpolated} increased clean accuracy of adversarially trained models by using augmentation methods \cite{zhang2017mixup,verma2018manifold}. Similarly, Zhang \etal \cite{zhang2020adversarial} also proposed to create adversaries using augmentation while updating the model on these adversaries using perturbed labels. However, these  methods \cite{lamb2019interpolated,zhang2017mixup,zhang2020adversarial} generate adversaries by mixing a sample with another which may or may not come from the same class, therefore the inter-class margins might not be enforced during training. In comparison, we carefully craft adversaries by style transfer followed by max-margin adversarial training, that results in enhanced robustness. Similarly, Xie \etal \cite{xie2020adversarial} introduce auxiliary batchnorm to keep adversarial statistics separate from the natural images, and Shu \etal \cite{shu2020preparing} adversarially perturb the features using adverarial batchnorm. In contrast, our method trains single batchnorm to jointly model statistics of adversarial and natural images and is capable of handling larger distortions. Another approach \cite{li2020shape} proposes to create shape and texture cue conflict during training. Our method introduces such conflict by injecting content and style information of target samples during the adversarial attack.

\begin{figure*}[htp]
  	\centering
    \includegraphics[width=0.8\linewidth]{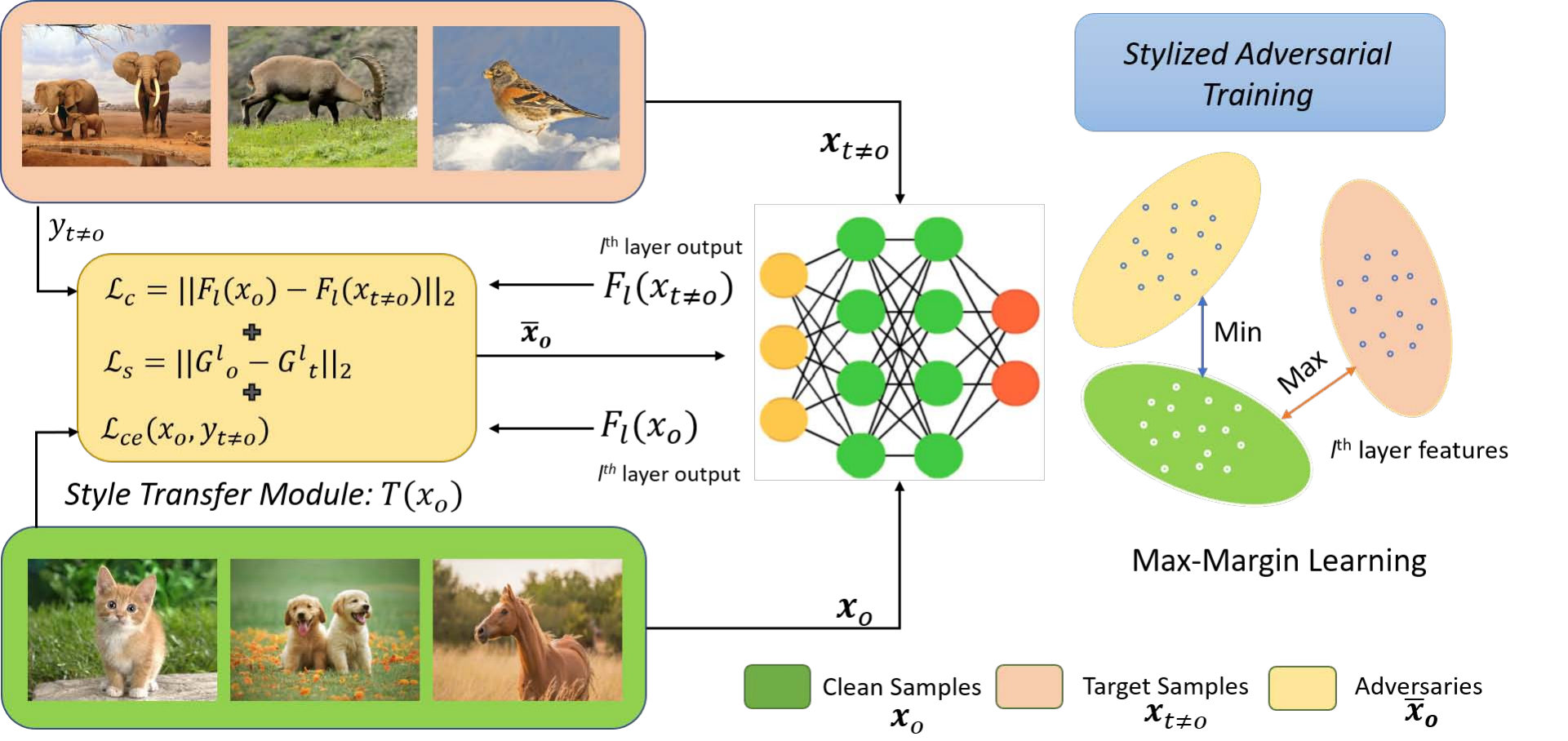}
      \caption{ \emph{Stylized Adversarial Training (SAT)}. Our style transfer module (\emph{left}) crafts perturbations based on three complimentary cues, that include content ($\mathcal{L}_{c}$) and style ($\mathcal{L}_{s}$) of the target image as well as the classifier boundary information ($\mathcal{L}_{ce}$). Based on the generated perturbations, our adversarial training approach seeks to minimize the distance between clean and adversarial examples of the same class and maximize the inter-class distances (\emph{right}).
  }
   \label{fig:method}
\end{figure*}

\noindent \textbf{Metric Learning Defenses:} More recently, some adversarial training efforts maximize the margin between clean and adversarial examples of different class samples. Mustafa \etal \cite{Mustafa_2019_ICCV} proposed a contrastive loss function to maximize the inter-class distances along the feature hierarchy of a deep network. \cite{ding2018max} dynamically selects the right perturbation budget for each data point to better enforce margin constraints during adversarial training. Triplet loss has also been explored to enforce margin constraints during training \cite{zhong2019adversarial,li2019improving,mao2019metric}. Similar to these approaches, our proposed method is model-agnostic and incorporates distance-based learning scheme. However, different from the previous works, we first transfer the style, content and boundary information from a target sample to the input and then maximize the distance between the target and the perturbed samples. In this manner, our triplet creation is automatic and does not need careful sample selection as in \cite{mao2019metric}.

\section{Methodology}\label{sec:methodoloy}
Consider a classifier, $\mathcal{F}(.)$ that maps input samples, $\bm{x} \in \mathbb{R}^{d}$, drawn from a dataset, $\bm{X}$, to a discriminative space $\mathcal{F}(\bm{x}) \in \mathbb{R}^{n}$, where $n$ represents the number of categories. Classifier can learn this mapping by minimizing an empirical risk defined on $\bm{X}$. Further, suppose that $\mathcal{F}_{l}(.)$ represents a feature map at the $l^{th}$ layer, and $\mathcal{T}$ denotes a transformation operation that keeps the output close to input i.e., $  \| \bm{x}_{o} - \mathcal{T}(\bm{x}_{o}) \| < \epsilon$, where $\epsilon$ is the perturbation budget. We present a generic training mechanism focused on robustifying neural networks by \emph{minimizing} feature difference between the original examples  $\bm{x}_{o}$ and the transformed positive samples $\mathcal{T}(\bm{x}_{o})$ and \emph{maximizing} feature difference between $\bm{x}_{o}$ and targeted class samples $\bm{x}_{t \neq o}$. The contrastive constraints can be achieved by minimizing the following loss function:
\begin{multline}\label{eq:constrastive_margin_loss}
    \mathcal{L}_{m}(\bm{x}_{o}, \bm{x}_{t \neq o}; \mathcal{T}, \mathcal{F}) = \text{max} \{\|\mathcal{F}_{l}(\bm{x}_{o}) -  \mathcal{F}_{l}(\mathcal{T}(\bm{x}_{o}))\|_p - \\ \|\mathcal{F}_{l}(\bm{x}_{o}) - \mathcal{F}_{l}(\bm{x}_{t \neq o})\|_p + m, 0 \},
\end{multline}
where $m$ represents the margin and $\| \cdot \|_p$ denotes $p$-norm. The transformation, $\mathcal{T}$, plays a significant role in training and should satisfy the following two properties:
\begin{itemize}
    \item The transformation maps output close to the input \ie, $\mathcal{T}(\bm{x}_{o}) \approx \bm{x}_{o}$.
    \item $\mathcal{T}$ should correlate with adversarial noise that fools the network.
\end{itemize}
Next in Sec.~\ref{sec:Transform}, we elaborate our proposed stylized perturbation generation mechanism ($\mathcal{T}$) that is central to our proposed defense described in Sec.~\ref{sec:SAT}.

\subsection{Transformation: Stylized Adversary Generation}\label{sec:Transform}
The choice of transformation $\mathcal{T}$ is critical to the strength of robustness achieved with adversarial training. Here, we present our transformation mechanism, achieved with a style transfer module (Fig.~\ref{fig:method}), that jointly utilizes the style, content and class-boundary information to craft deceptive perturbations. The overall objective for learning the adversarial transformation is, 
\begin{align}
    \label{eq:adversarial_transformation}
    \underset{\mathcal{T}}{\text{argmax}} \;\; & \mathcal{L} \left( \mathcal{F}\left(\mathcal{T}(\bm{x}_{o})\right),  \mathcal{F}(\bm{x}_{o}) \right) , \notag\\
    &\text{s.t.,} \; \|\mathcal{T}(\bm{x}_{o}) - \bm{x}_{o}\|_{\infty} \le \epsilon ,
\end{align}
where, $\mathcal{L}$ denotes any loss function and $\epsilon$ is the allowed perturbation budget. The aim is to remain in the vicinity of input sample $\bm{x}_{o}$, but maximally alter the predicted output by the model $\mathcal{F}$.  The above objective is pursued in previous adversary generation methods as well, however, our main difference is the way we incorporate target samples ($\bm{x}_{t\neq 0}$) while crafting adversaries. Specifically, we extract three types of information about the target sample including class-boundary information, image style and visual content. 
For example, in an effort to robustify the model, the adversarial transformation should create adversarial examples that contain style and texture of the samples from target classes $\bm{x}_{t\neq 0}$, within a given perturbation budget, $\epsilon$. The following adversarial loss is minimized to learn the transformation $\mathcal{T}$:
\begin{equation}\label{eq:adv_generation_loss}
    \mathcal{L}_{adv} =
    \underbrace{\alpha\cdot\mathcal{L}_{s}}_{Style\:loss}  + \underbrace{\gamma\cdot\mathcal{L}_{c}}_{Content\:loss}+  \underbrace{\beta \cdot\mathcal{L}_{ce}}_{Cross-entropy \:loss},
\end{equation}
where $\alpha, \gamma, \beta$ denote the hyper-parameters used for loss re-weighting which are set via validation. Notably, the style and content loss components are computed within feature space while boundary information comes form the logit space. We explain the individual losses below. 

\noindent\textbf{Style loss:}
The objective of the style loss $\mathcal{L}_{s}$ is to transfer the texture of the target image to fool the classifier, $\mathcal{F}$. Style transfer \cite{gatys2016image} can be achieved by minimizing the mean-squared distance between the Gram matrices obtained from the feature maps at layer $l$ of the original and targeted images, as follows:
\begin{equation}
    \mathcal{L}_{s} = \|\bm{G}_{o}^{l} - \bm{G}_{t\neq o}^{l}\|_2^2, \quad \text{s.t., } \bm{G}^l = \bm{f} \bm{f}^{T},
\end{equation}
where $\bm{G} \in \mathbb{R}^{c\times c}$ represents the Gram matrix, $\bm{f} \in \mathbb{R}^{c \times (h.w)}$ denotes the matrix formed by stacking together the channel-wise features from $\mathcal{F}_{l}(\bm{x}) \in \mathbb{R}^{c \times h \times w}$. Here, $h, w$ and $c$ denote the height, width and channel dimensions of the feature tensor from layer $l$, respectively. 

\noindent\textbf{Content loss:}
The objective of the content loss $\mathcal{L}_{c}$ is to mix the content of the target class image with that of the original image. This is achieved by minimizing the mean-squared distance between the feature representations of  $\bm{x}_{o}$ and  $\bm{x}_{t\neq o}$, as follows:
\begin{equation}
    \mathcal{L}_{c} = \|\mathcal{F}_{l}(\bm{x}_{o}) - \mathcal{F}_{l}(\bm{x}_{t\neq o})\|_2^2.
\end{equation}

\noindent\textbf{Boundary loss:}
The objective of this loss is to push the transformed sample into the boundary of targeted class. For this purpose, we use the regular cross-entropy loss. If $\bm{y}_{t}$ represents target label then boundary-based targeted attack can be achieved by minimizing $\mathcal{L}_{ce}(\bm{x}_{o}, \bm{y}_t)$. 

\subsection{Stylized Adversarial Training}\label{sec:SAT}
We describe our robust training framework in Algorithm \ref{alg:rtf} that apply adversarial transformation to push clean samples $\bm{x}_{o}$ toward the targeted samples $\bm{x}_{t\neq 0}$ within a predefined budget $\epsilon$, and then robustify the network with adversarial training. The adversarial training procedure employs cross-entropy loss alongside the contrastive loss $\mathcal{L}_m$ that seeks to maximize inter-class margins (Eq. \ref{eq:constrastive_margin_loss}).

\begin{algorithm}[t]
\small
\caption{SAT: Stylized Adversarial Training}
\label{alg:rtf}
\begin{algorithmic}[1]
\Require A classifier $\mathcal{F}$, clean sample $\bm{x}_{o}$ and their corresponding labels $\bm{y}$, targeted sample $\bm{x}_{t \neq o}$ and their corresponding labels $\bm{y}_{t}$, margin loss $\mathcal{L}_{m}$, cross-entropy loss $\mathcal{L}_{ce}$, $w_1$, $w_2$, no. of attack steps $n$, step size $\xi$, and no. of iterations $T$.
\For {$t = 1$ to $T$}
\State Initialize: $\bm{\bar{x}}_{o} = \bm{x}_{o}$
\For {$i=1$ to $n$} 
\State Forward pass $\bm{\bar{x}}_{o}$, and $\bm{x}_{t \neq o}$ to $\mathcal{F}$ and compute adversarial loss $\mathcal{L}_{adv}$ (Eq. \ref{eq:adv_generation_loss});
\State Compute gradient noise, $\bm{g}_{t} = \nabla_{\bm{x}} \,\mathcal{L}_{adv}$;
\State Generate adversaries using;
\begin{equation}
\bm{\bar{x}}_{o} = \text{Clip} \left( \bm{\bar{x}}_{o} -\xi\cdot\mathrm{sign}(\bm{g}_{t}), \bm{x}_{o}+\epsilon, \bm{x}_{o}-\epsilon \right);
\end{equation}
\EndFor
\State Forward pass $\bm{\bar{x}}_{o}$ through $\mathcal{F}$;
\State Backpass and update the parameters of $\mathcal{F}$ to minimize the combined loss: 
\begin{equation}\label{eq:overall_loss}
    \mathcal{L} = w_{1}\cdot\mathcal{L}_{m}(\bm{x}, \bm{\bar{x}}_{o}, \bm{x}_{t \neq o}) + w_{2}\cdot\mathcal{L}_{ce}(\bm{\bar{x}}_{o}, \bm{y})
\end{equation}
\EndFor \\
\Return Robust classifier, $\mathcal{F}$.
\end{algorithmic}
\end{algorithm}

\textbf{Non-Adversarial $\mathcal{T}$:} In order to emphasize on the significance of transformation $\mathcal{T}$ in the SAT framework, here we consider a non-adversarial transformation function.  In this case, $\mathcal{T}$ can be  as simple as adding Gaussian noise to the clean samples. Such a transformation is computationally less expensive and has been studied before in \cite{cohen2019certified,Kannan2018AdversarialLP,zantedeschi2017efficient}. In this case, the perturbation generation process in lines 2-4 of Algorithm~\ref{alg:rtf} is simply replaced with adding  randomly sampled Gaussian noise in the image. We explore the effect of non-adversarial transformation on adversarial robustness and compare our training Algorithm \ref{alg:rtf} with a recently proposed `Guided Cross-Entropy' (GCE) \cite{chen2019improving} method in Sec.~\ref{subsec:results_abstract_training_framework}. We note that in the non-adversarial scenario, targeted samples are not playing any significant role other than providing a reference to contrastive margin loss which leads to a sub-optimal solution (see Sec.~\ref{subsec:results_abstract_training_framework}). This shows the efficacy of proposed stylized perturbation generation approach (Sec.~\ref{sec:Transform}).

\begin{table*}[!htp]
	\centering\setlength{\tabcolsep}{7pt}
		\resizebox{1.8\columnwidth}{!}{
		\begin{tabular}{l|l|c|cccccccc}
			\toprule
			\multirow{2}{*}{Model}&\multirow{2}{*}{Defense}&\multirow{2}{*}{Clean} & \multicolumn{2}{c}{PGD\cite{madry2018towards}}&CW \cite{feature_scatter}&MIFGSM \cite{dong2018boosting}&DeepFool\cite{moosavi2016deepfool}& ODI \cite{tashiro2020ods} & SPSA \cite{uesato2018adversarial} & AA \cite{croce2020reliable} \\
			\cline{4-11}
			& & &  {20} & {1000}&100&100&100 \Tstrut\\
			\hline
			\multirow{4}{*}{ResNet18} & NT & 94.5&0.0&0.0&0.0&0.0&1.2&0.0&0.0&0.0\\
			&  Trades \cite{Zhang2019theoretically} ($\lambda = 1$) & 91.3 & 26.5&--&--&--&--&--&--&--\\
			&  Trades \cite{Zhang2019theoretically} ($\lambda = 5$) & 81.7 & 50.6&50.1&49.3&52.1&63.0&--&--&--\\
			\cline{2-11}
			&  SAT ($n=1$) &92.1& 69.1 & 66.8 & 62.5 & 72.8 & 66.0& 59.9& 21.2 & 10.8 \Tstrut\\
			& SAT ($n=10$) & 90.0 & 51.3 & 51.0 & 50.6 & 56.7 & 64.1 & 55.5 & 52.4 & 50.2  \\
			\hline
	\multirow{4}{*}{WideResNet} & NT & 95.3 &0.0&0.0&0.0&0.0&3.2 & 0.0& 0.0 & 0.0\Tstrut\\
			&  Trades \cite{Zhang2019theoretically} ($\lambda = 1$) & 88.6 & 49.1&48.9&48.3&53.5&69.0 &--&--&--\\
			&  Trades \cite{Zhang2019theoretically} ($\lambda = 6$) & 84.9& 56.6&56.4&53.7&58.5&72.0& 54.5 & 54.7 &53.1\\
			\cline{2-11}
			&  SAT ($n=1$) &93.3 & 80.7 & 72.2 & 71.8 & 82.8 & 88.3 & 69.2 & 23.1& 13.4 \Tstrut\\
			&  SAT ($n=10$) & 91.3 & 55.5&55.1 &55.4 & 59.8 & 73.4 & 57.9 & 55.8 & 53.9\\
			\bottomrule
		\end{tabular}
	}
	\vspace{0.2em}
	\caption{\textbf{White-box} attack scenario. Comparisons of our defense with Trades \cite{Zhang2019theoretically} on CIFAR10 test set under perturbation budget $\epsilon \le 8$. Models trained via our proposed approach (Algorithm \ref{alg:rtf}) not only withstand PGD attack with 1000 iteration but also Auto-Attack as well, while providing high accuracy on clean samples. We used DeepFool  with default settings and project the adversarial noise found by the attack  on $l_\infty$ with $\epsilon \le 8$.}
	\label{tab:comparison_with_trades}
\end{table*}


\begin{figure*}[t]
\centering
  \begin{minipage}{.33\textwidth}
  	\centering
    \includegraphics[  width=\linewidth, keepaspectratio]{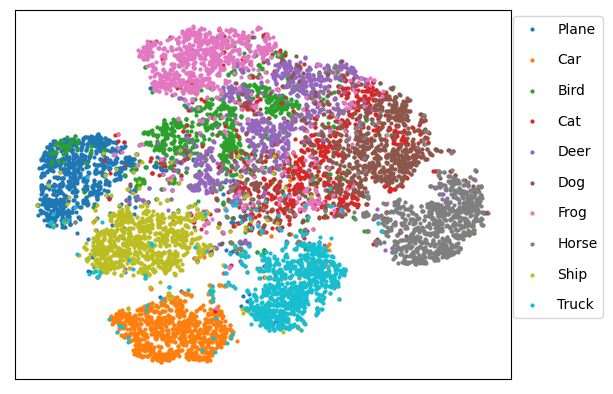}
  \end{minipage} 
  \begin{minipage}{.33\textwidth}
  	\centering
    \includegraphics[width=\linewidth, keepaspectratio]{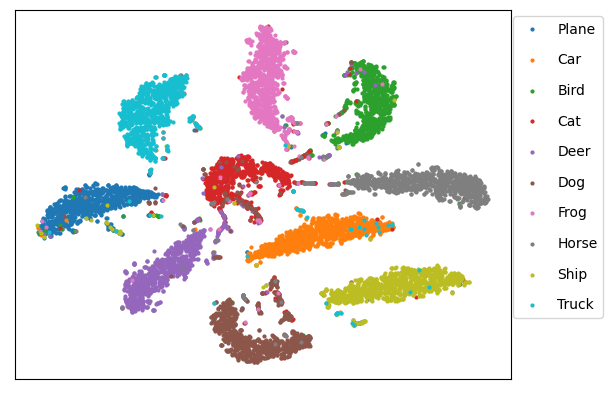}
  \end{minipage} 
   \begin{minipage}{.33\textwidth}
  	\centering
    \includegraphics[width=\linewidth, keepaspectratio]{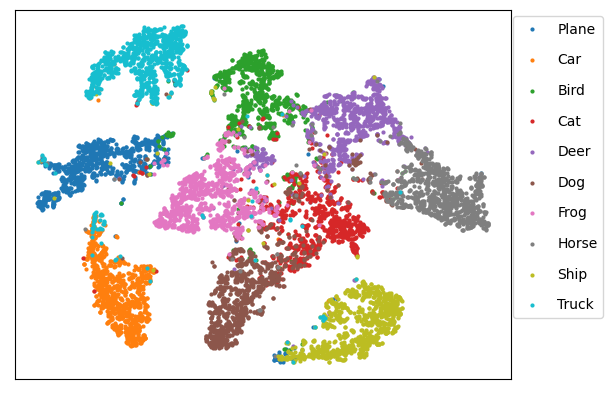}
  \end{minipage} 
  \caption{\centering (Left to right) Latent space t-SNE visualization of the logits extracted from Trades \cite{Zhang2019theoretically}, our SAT  ($n=1$) and our SAT ($n=10$) on CIFAR10 test set. Compared
to Trades \cite{Zhang2019theoretically}, our SAT models form distinct class-wise clusters which help retain clean accuracy and improves robustness. }
\label{fig:tsne}
\end{figure*}

\begin{table*}[t]
	\centering\setlength{\tabcolsep}{5pt}
		\resizebox{0.85\linewidth}{!}{
		\begin{tabular}{ l|c|c|cccccc|cccccc|c}
			\toprule
			\multirow{2}{*}{Defense}&\multirow{2}{*}{Clean}& \multirow{2}{*}{FGSM} & \multicolumn{6}{c}{PGD \cite{madry2018towards}}&\multicolumn{6}{c}{CW \cite{feature_scatter}} & AA \cite{croce2020reliable}\\
			\cline{4-16}
			& &  &   10 & 20 & 40  & 100 & 500 & 1000 & 10 & 20 & 40 & 100& 500&1000 \Tstrut\\
			\hline
			NT & \textbf{95.6} & 36.9 &  0.0 & 0.0& 0.0 & 0.0  & 0.0 & 0.0 & 0.0 & 0.0 & 0.0 & 0.0&0.0&0.0 & 0.0\\
			
			PAT \cite{madry2018towards} & 85.7 & 54.9  & 45.1 & 44.9 & 44.8 & 44.8 & 44.8 & 44.6 & 45.9 & 45.7&  45.6& 45.4 & 45.2 & 45.1 & 44.0\\   
			BL \cite{wang2019bilateral} & 91.2&  70.7  &  -- & 57.5 &  -- & 55.2 &--&-- &--& 56.2  & -- & 53.8 & -- & -&--\\ 
			FS \cite{feature_scatter}& 90.0& 78.4  & 70.9 & 70.5  & 70.3 &  68.6 &62.8&61.0& 62.6 & 62.4  & 62.1 &  60.6 & 58.2 & 58.1 & 36.6\\ 
			\hline
			SAT ($n=1$) & 93.3 &85.0&81.1&80.7&79.8&78.5&73.7&72.1&75.0&74.9&73.2&71.8 & 71.2 & 71.0 & 13.4 \Tstrut\\
			SAT ($n=10$) & 91.3 & 63.7 & 55.7 & 55.5 & 55.3 & 55.3 & 55.1 & 55.1 & 55.7 & 55.6 & 55.4 & 55.2 & 55.2 & 55.1 & 53.9 \Tstrut\\
			\bottomrule
		\end{tabular}
	}
	\vspace{0.2em}
	\caption{\textbf{White-box} attack scenario. Comparison (\%) of our approach with naturally trained (NT), Madry (PAT) \cite{madry2018towards},  bilateral (BL) \cite{wang2019bilateral} and feature scattering (FS) \cite{feature_scatter} methods on CIFAR10 test set under different threat models. Attacks ran for maximum of 1000 iterations. Models trained using our approach (Algorithm \ref{alg:rtf})  show significant robustness regardless of attack iterations or attack type while providing noticeable clean accuracy.}
	\label{tab:cifar10_comparision_with_fs}
\end{table*}

\begin{table*}[!htb]
\begin{minipage}{0.5\textwidth}
      \centering
      \scalebox{0.9}{
  	\begin{tabular}{ l|c|c|cc|cc|c}
			\toprule
			\multirow{2}{*}{Defense}&\multirow{2}{*}{Clean}& \multirow{2}{*}{FGSM} & \multicolumn{2}{c}{PGD \cite{madry2018towards}}&\multicolumn{2}{c}{CW \cite{feature_scatter}} & AA \cite{croce2020reliable}\\
			\cline{4-8}
			& &  &     {20} &  {100} &   {20} &  {100} \Tstrut\\
			\hline
			NT &\textbf{97.2} & 53.0  & 0.3 & 0.1& 0.3 & 0.1 & 0.0   \\
			
			PAT \cite{madry2018towards} & 93.9 &	68.4 &	47.9 & 46.0 & 48.7 & 47.3 & --\\   
			BL \cite{wang2019bilateral} & 94.1  &69.8  &53.9 & 50.3 & -- & 48.9 & -- \\ 
			FS \cite{feature_scatter}& 96.2& 83.5 & 62.9  & 52.0 & 61.3 & 50.8 & 37.8 \\ 
			\hline
			SAT ($n=1$) & 96.2 & 86.0&73.2&71.5&70.0&68.0 & 27.9 \Tstrut\\
			SAT ($n=10$) & 95.5 & 75.6 & 62.5 & 62.0 & 63.3 & 63.1 & 58.4 \\
			\bottomrule
  \end{tabular}}
  \end{minipage}
\begin{minipage}{.50\textwidth}
          \centering
    \scalebox{0.9}{
    \begin{tabular}{ l|c|c|cc|cc|c}
			\toprule
			\multirow{2}{*}{Defense}&\multirow{2}{*}{Clean}& \multirow{2}{*}{FGSM} & \multicolumn{2}{c}{PGD \cite{madry2018towards}}&\multicolumn{2}{c}{CW \cite{feature_scatter}} & AA \cite{croce2020reliable}\\
			\cline{4-8}
			& &  &     {20} &  {100} &   {20} &  {100} \Tstrut\\
			\hline
			NT & \textbf{79.0} &  10.0 & 0.0 & 0.0 & 0.0 & 0.0 & 0.0\\
			
			PAT \cite{madry2018towards} & 59.9 &  28.5 &  22.6 & 22.3 & 23.2 & 23.0 & 21.1\\   
			BL \cite{wang2019bilateral} & 68.2 & 60.8 & 26.7 & 25.3&  -- & 22.1 & --\\ 
			FS \cite{feature_scatter}&  73.9& 61.0 & 47.2 & 46.2 & 34.6 &  30.6& 1.1 \\ 
			\hline
			SAT ($n=1$) & 74.1 & 64.9 & 49.7 & 49.1 & 44.2 & 40.6 & 4.7\Tstrut\\
			SAT ($n=10$) & 70.1 &34.5&26.7&26.4&26.8&26.5&\textbf{25.6} \\
			\bottomrule
  \end{tabular}}
\end{minipage}
\vspace{0.2em}
\caption{\textbf{White-box} attack scenario. Comparison (\%) is shown on SVHN (left) and CIFAR100 (right) test sets.}
\label{tab:cifar100_comparision_with_fs}
\vspace{-0.5em}
\end{table*}

\section{Experimental Protocol}
We experiment on the widely used SVHN \cite{netzer2011reading}, CIFAR10, and CIFAR100 datasets \cite{cifar}. We show comparative studies on the ResNet18, WideResNet  and ResNet101 models. 
We set $n=1, \xi=\epsilon$ and $n=10, \xi=\frac{2}{255}$ for single-step and multi-step SAT, respectively (Algorithm \ref{alg:rtf}). The pixel values are normalized within [-1,+1] before forward-pass and label smoothing \cite{feature_scatter} is used during training. Unless otherwise mentioned, the perturbation budget $\epsilon$ is set to 8 (out of 255). NT and AT respectively denote naturally and adversarially trained models. All hyper-parameters are tuned for single-step SAT (Sec. \ref{sec:ablations}). Specifically, we set $\alpha=\gamma=\beta=1$ in Eq. \ref{eq:adv_generation_loss} and $w_1=w_2=1$ in Eq. \ref{eq:overall_loss}. All models are trained for 200 epochs with margin and $p$-norm in Eq. \ref{eq:constrastive_margin_loss} set to 1 and 2, respectively. \textbf{\emph{Reliable Adversarial Evaluation:}} Our models are evaluated against PGD attack proposed by \cite{madry2018towards} and CW which is a variant of PGD attack with margin based loss \cite{carlini2017towards},  MIFGSM \cite{dong2018boosting}, DeepFool \cite{moosavi2016deepfool}, ODI (PGD variant) \cite{tashiro2020ods}, SPSA \cite{uesato2018adversarial}, and Auto-Attack \cite{croce2020reliable}. We used these attacks with default settings. We used cleverhans \cite{papernot2018cleverhans} implementation of SPSA and open-source implementations of ODI, and Auto-Attack. We used Auto-Attack with standard setting which is an ensemble of attacks with random restarts including boundary based attack FAB \cite{croce2020minimally} and query based attack Square \cite{ACFH2020square} with no less than 5000 queries.  Our code and pretrained models are available at \url{https://github.com/Muzammal-Naseer/SAT}.

\subsection{SAT: Defense Results and Insights}\label{subsec:results_robust_training_framework}
We thoroughly investigate the effect of our proposed adversarial transformation (Sec.~\ref{sec:methodoloy}) to maximize adversarial robustness without compromising clean accuracy. Our analysis is divided into the four categories:
\textbf{(a)} Robustness against constrained adversarial attacks  ($l_\infty \le 8$),
\textbf{(b)} Robustness against unconstrained adversarial attacks, namely Rectangular Occlusion Attacks (ROA) \cite{Wu2020Defending}, which completely destroy the image content within a given window size,
\textbf{(c)} Robustness against natural distributional shifts in data, and
\textbf{(d)} Robustness against common corruptions.

\subsubsection{Robustness against constrained adversarial attacks} We compare our approach with the state-of-the-art methods including Trades \cite{Zhang2019theoretically}, Feature Scattering (FS) \cite{feature_scatter} and the metric learning based Prototype Conformity Loss (PCL) \cite{Mustafa_2019_ICCV}. For a fair and direct comparison, we follow the same threat models (attack settings) and network architectures as recommended in the papers of the respective methods. Further, we compare the performance under ensemble of attacks (Auto-Attack \cite{croce2020reliable}) for reliable adversarial evaluation under constrained attack setting ($\epsilon \le 8$).

\noindent\textit{\textbf{Comparison with Trades \cite{Zhang2019theoretically}}} is presented in Table \ref{tab:comparison_with_trades}. We note that \cite{Zhang2019theoretically} offers a trade-off parameter ($\lambda$) to increase robustness on the cost of losing clean accuracy. Our defense SAT ($n=10$) not only achieves 0.8 higher robustness in top-1 accuracy when compared with best adversarial results from \cite{Zhang2019theoretically} ($\lambda=6$) against Auto-Attack but also improves clean accuracy by 6.4. 
Furthermore, our defense can withstand large number of PGD attack iterations \eg the drop in robustness (top-1 accuracy) is only 0.6 when attack iterations increase from 20 to 1000 on our SAT model (Table \ref{tab:cifar10_comparision_with_fs}). As a result of our proposed max-margin learning, class-wise latent features of our defense are well clustered and distinctly separated as compared to \cite{Zhang2019theoretically} (see Figure \ref{fig:tsne}). This leads to not only better robustness against major attacks including PGD, CW, MIFGSM, DeepFool and Auto-Attack but also higher generalization on common corruptions (Table \ref{tab:results_on_common_corruptions}).

\begin{table*}[t]
  	\centering\setlength{\tabcolsep}{10pt}
    \scalebox{1.0}{
	\begin{tabular}{ l|c|cccc|cccc}
		\toprule
		\multirow{2}{*}{Defense}&\multirow{2}{*}{Clean} & \multicolumn{4}{c}{ROA (Gradient Search)}&\multicolumn{4}{c}{ROA (Exhaustive Search)}\\
		\cline{3-10}
		&  &  5x5 & 7x7 & 9x9 & 11x11 & 5x5 & 7x7 & 9x9&11x11   \Tstrut\\
		\hline
		NT &96.0&56.0&42.3&27.3&13.0&38.5&21.0&8.6&2.4\\
		
		PAT \cite{madry2018towards} &86.8&47.2&29.7&15.4&7.0&33.8&23.7&7.9&3.0 \\  
		Trades \cite{Zhang2019theoretically} &84.9&49.8&30.5&16.1&7.1&39.2&28.0&8.8&3.2\\
		FS \cite{feature_scatter}& 90.0&66.5&56.2&44.3&32.6&49.1&42.2&22.3&12.5\ \\ 
		\hline
		SAT ($n=1$) & 93.3&\textbf{74.7}&\textbf{65.4}&\textbf{54.2}&\textbf{42.6}&\textbf{56.4}&\textbf{51.0}&\textbf{31.1}&\textbf{23.1} \Tstrut\\
		\bottomrule
	\end{tabular}
}
\vspace{0.2em}
  \caption{Adversarial robustness against unconstrained adversarial attack, ROA \cite{Wu2020Defending} at different window sizes. Our defense performs significantly better than other training approaches.}
\label{tab:cifar10_comparision_against_ROA}

\end{table*}

\begin{table}[t]
	\centering\setlength{\tabcolsep}{5pt}
		\resizebox{1\linewidth}{!}{
		\begin{tabular}{ l|c|c|cc|cc}
		\toprule
		 & \multicolumn{6}{c}{Targeted Attacks} \\ 
		 \cline{2-7}
		Defense& SA (ours) & FGSM & \multicolumn{2}{c}{PGD}&\multicolumn{2}{c}{CW}\\
		\cline{4-7}
		& &   & 5 &10 & 5 &10 \\
		\hline
		NT  & 52.3 & 27.5 & 97.0 & 98.1&97.8&99.0\\
		PAT \cite{madry2018towards} & 24.5& 19.8 & 23.1 &27.6&23.6&28.3\\
		Trades \cite{Zhang2019theoretically} & 20.0 & 16.6 & 19.8 &21.0& 20.3&22.9\\
		FS \cite{feature_scatter} & 19.4& 14.2 &20.7 & 24.2 & 21.3&25.4\\
		\hline
		SAT ($n=1$) & 12.5& 9.8 & 14.2& 16.4 &15.3&16.5  \\
		SAT ($n=10$) &  11.0 & 10.1 & 14.1 &13.9& 12.3&14.7 \\
		\bottomrule
		\end{tabular}
	}
	\vspace{0.2em}
	\caption{\textbf{White-box} target-attack scenario. Target attack success rate (\%) is reported on CIFAR10 test set. Our (single step) attack not only performs better than FGSM but also has competitive success rate compared to 5-step PGD and CW attacks.}
	\label{tab:stylized_attack_comparision}
\end{table}

\begin{table}
\setlength{\tabcolsep}{4pt}
\resizebox{\columnwidth}{!}{
\begin{tabular}{l|c|cccccc}
\multicolumn{7}{c}{(a) \textbf{CIFAR10.} Perturbation budget is 8/255 in $\ell_{\infty}$ norm.} \Tstrut \\
\toprule
Defense & Clean & FGSM & IFGSM  &  CW & MIFGSM &  PGD & AA \\
\hline
PCL \cite{Mustafa_2019_ICCV} & 91.9 & 74.9 & 46.0 & 51.8 & 49.3 & 46.7 & 11.2\\
SAT $(n=1)$ & 92.3 & 84.7 & 83.5 & 81.2 & 83.8 &83.5 & 11.1 \\
SAT $(n=10)$ & 90.1 & 54.7 & 55.2 & 55.1 & 56.0 & 54.7 & \textbf{50.1} \\
\bottomrule
\multicolumn{7}{c}{(b) \textbf{CIFAR100.} Perturbation budget is 8/255 in $\ell_{\infty}$ norm.} \Tstrut\\
\toprule
Defense &Clean & FGSM & IFGSM  &  CW & MIFGSM &  PGD \\
\hline
PCL \cite{Mustafa_2019_ICCV} & 68.3 & 60.9 & 34.1 & 36.7& 33.7 & 36.1 & 5.0\\
SAT $(n=1)$ &72.5 &65.2&48.3&47.5&49.2&48.0 & 3.3\\
SAT $(n=10)$ & 69.5 & 36.7 & 24.8 & 25.1 & 26.0 & 24.4 & \textbf{22.9} \\
\bottomrule
\multicolumn{7}{c}{(c) \textbf{SVHN.} Perturbation budget is 8/255 in $\ell_{\infty}$ norm.} \Tstrut \\
\toprule
Defense &Clean & FGSM & IFGSM  &  CW & MIFGSM &  PGD \\
\hline
PCL \cite{Mustafa_2019_ICCV} & 94.4 & 76.5 & 48.8 & 54.8 & 47.1 & 47.7 & 14.4\\
SAT $(n=1)$ & 95.3&85.7&67.8&65.3&68.0&66.8 & 20.6\\
SAT $(n=10)$ & 94.1 & 78.0 & 52.3 & 52.1& 55.8 & 51.8 & \textbf{49.5}\\
\bottomrule
\end{tabular} }
\vspace{0.2em}
\caption{\textbf{White-box:} Comparison between PCL and our proposed defense (SAT) under the threat model of PCL. SAT shows significantly better robustness as compared to PCL. We train and evaluate SAT on the ResNet models used by PCL  for consistent comparisons.}
\label{tab:comparison_with_pcl}
\end{table}

\noindent\textit{\textbf{Comparison with FS \cite{feature_scatter}}} is presented in Tables \ref{tab:cifar10_comparision_with_fs} and \ref{tab:cifar100_comparision_with_fs}. In terms of worst-case robustness measure (Auto-Attack which is an ensemble of attacks including PGD with different loss objectives with multiple random restarts, boundary based attack FAB \cite{croce2020minimally} and query based square attack \cite{ACFH2020square}), our defense offers 17.3, 24.5 and 20.6 (Tables \ref{tab:cifar10_comparision_with_fs} and \ref{tab:cifar100_comparision_with_fs}) robustness gain in top-1 accuracy on CIFAR10, CIFAR100 and SVHN datasets, respectively. We further observe that our approach simultaneously improves clean accuracy while achieving significant robustness gains. In contrast to SAT (Algorithm \ref{alg:rtf}), FS \cite{feature_scatter} is dependant on optimal transport distance to increase model loss towards the unknown class. It does not leverage target image information and neither it offers any mechanism to increase inter-class margins.

\noindent\textit{\textbf{Comparison with PCL \cite{Mustafa_2019_ICCV}:}} The results in Table \ref{tab:comparison_with_pcl} indicate that our defense demonstrates significant robustness gains as compared to metric-learning based prototype conformity loss (PCL) \cite{Mustafa_2019_ICCV}, boosting adversarial accuracy by 38.9, 17.9 and 35.1 on CIFAR10, CIFAR100 and SVHN, respectively. We further observe that alongside significant gains for adversarial robustness, our approach also increases the clean accuracy on all the evaluated datasets. We note that PCL \cite{Mustafa_2019_ICCV} enhances separation between class centers and is dependent on PGD untargetted attack, while SAT increases distance between the features of original and target samples, which are also responsible for adversarial perturbations.

\textbf{Effect of Stylized Attack Steps on SAT:}
We study the effect of the number of steps of our proposed stylized attack on adversarial training. We set $n=1$ for our single-step and $n=10$ for multi-step SAT (Algorithm \ref{alg:rtf}) following baseline adversarial training methods \cite{goodfellow2014explaining, madry2018towards, Zhang2019theoretically}. We observe that single-step SAT is computationally efficient and provides better generalization against common corruptions than naturally trained or other single-step adversarial training methods such feature scattering \cite{feature_scatter} (Table \ref{tab:results_on_common_corruptions}). However, single-step SAT has variable performance under different threat models (attack settings). On the other hand, multi-step SAT is computationally expensive but perform consistently against different attacks including an ensemble of attacks, Auto-Attack \cite{croce2020reliable}, (Tables \ref{tab:comparison_with_trades}, \ref{tab:cifar100_comparision_with_fs} \& \ref{tab:comparison_with_pcl}).

\subsubsection{Robustness against unconstrained attack}
Here, we consider the unconstrained occlusion attack called ROA \cite{Wu2020Defending} against our proposed adversarial training approach. We run the gradient and exhaustive search versions of ROA on CIFAR10 dataset with four different window sizes (ranging from 5x5 to 11x11).
We observe that as the window size is increased for ROA, the robustness of PAT and Trades is matched with a model trained without adversarial training (NT). In comparison,  our defense shows a higher robustness e.g., an relative increment of 84\% and 620\% more over FS and Trades respectively, at the window size of 11x11 (see Table \ref{tab:cifar10_comparision_against_ROA}). This is attributed to the fact that our training approach constructs a smooth loss surface and requires large input distortions to deceive the model. We empirically demonstrate such smoothness by analyzing the intermediate features of PAT \cite{madry2018towards} and SAT in terms of correlation loss (CL) between features of adversarial image with respect to the clean image. Lower the correlation, better the feature space as it indicates that model feature space does not change significantly in response to the attack. We ran 100 iterations of PGD attack and extract features from the last layer before the logit layer. Correlation loss is measured in terms Frobenius norm between covariance of adversarial and clean features. The averaged  correlation loss of our method is 0.74 as compared to 8.8 from PAT \cite{madry2018towards} on the CIFAR10 dataset. This further supports the robustness of our proposed adversarial training approach.

\subsubsection{Robustness against stylized attack}
We evaluate different defenses on our proposed single-step stylized attack (SA). Stylized attack (SA) is targeted in nature. We measure the robustness of different adversarial training mechanisms against the targeted attacks in Table~\ref{tab:stylized_attack_comparision}. Given a sample of a certain source class in a test set of $N$ classes, we adversarially optimized it towards the other $N-1$ target classes. This process is repeated for all the samples in a test set.  We consider the attack as successful if the model misclassifies  an adversarial sample as the target attack class. We report the average target attack success rate across N targets (\eg, 10 in the case of CIFAR10). Our attack performs better than computationally similar FGSM (\eg, 90\% higher success rate against naturally trained model) and on-par with 5-steps PGD and CW attacks against adversarially trained models (Table~\ref{tab:stylized_attack_comparision}). We observed that success rate of stylized attack against SAT is significantly low as compared to the baseline defenses. For example, success rate of SA against SAT is 36\% lower than FS \cite{feature_scatter}. This is due to the fact that SAT is trained on similar adversaries generated by SA during adversarial training. However, this does not lead to overfitting as for the case of models trained via single step FGSM \cite{goodfellow2014explaining}. In fact our model generalizes to other target and untargeted attacks (Tables \ref{tab:cifar10_comparision_with_fs}-\ref{tab:stylized_attack_comparision}). This is contributed to our novel adversarial training method which takes the behavior of adversarial attack into account while updating the model weights.

\subsubsection{Robustness against natural distributional shifts} 
\begin{figure}[t]
    \centering
  \begin{minipage}{.49\columnwidth}
  	\centering
    \includegraphics[  width=\linewidth, keepaspectratio]{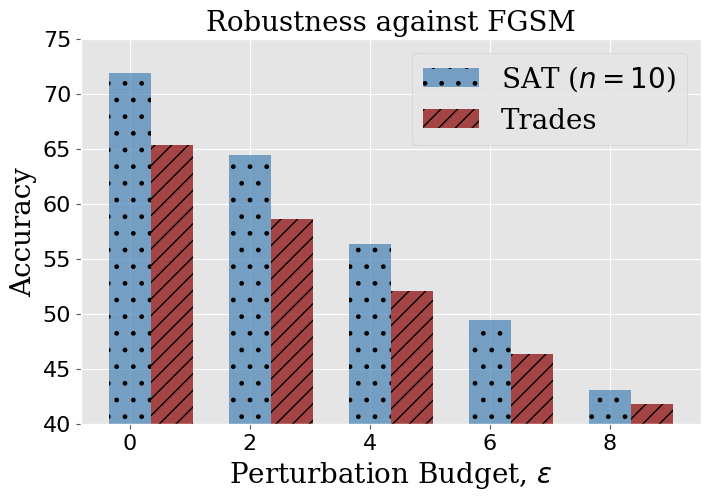}
  \end{minipage} 
  \begin{minipage}{.49\columnwidth}
  	\centering
    \includegraphics[width=\linewidth, keepaspectratio]{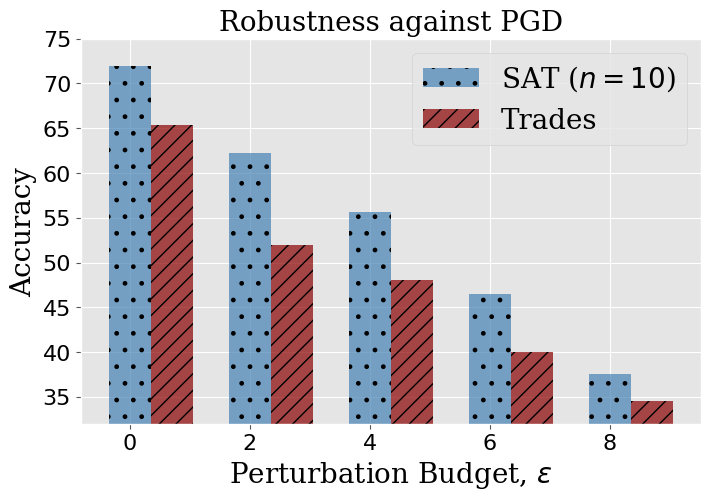}
  \end{minipage} 
  \caption{\textbf{White-Box analysis (\%):} Our proposed SAT handles the data distributional shifts significantly better than Trades \cite{Zhang2019theoretically}. Models are evaluated on CINIC-10 \cite{Darlow2018CINIC10IN} test set. PGD attack ran for 10 iterations. SAT's state-of-the-art robustness to such distributional shifts complements the strength of our proposed approach to enhance the generalizability of deep neural networks. }
\label{fig:robustness_to_distibutional_shift}
\end{figure}

Generalization of deep  networks goes beyond adversarial robustness \eg robustness to non-adversarial distributional shifts. For example, \cite{recht2018cifar} showed that models trained on CIFAR10 suffer from accuracy drop when there are small natural distributional shifts in the data. We evaluated robustness of adversarially trained models on CINIC-10 test set (90k images). CINIC-10 is down-sampled from ImageNet \cite{ILSVRC15} and contains the same classes as in CIFAR10 but with a significantly different distribution. Fig.~\ref{fig:robustness_to_distibutional_shift} shows that SAT outperforms Trades \cite{Zhang2019theoretically} and generalizes well to the shift in underlying data distribution. This is potentially because the SAT simulates distribution shifts in style during its training.

\begin{table}[t]
	\centering\setlength{\tabcolsep}{4pt}
			\scalebox{1.1}{
			\begin{tabular}{lcccc}
				\toprule
				Corruption & NT  & Trades \cite{Zhang2019theoretically} & FS \cite{feature_scatter} & SAT ($n=1$)\\
				\midrule
				Brightness& 93.7 &80.6&88.3&92.1\\
				Contrast& 92.1 & 43.1&82.5&88.9 \\
                Defocus Blur& 92.3  &80.0&86.1&90.4 \\
                Elastic Transform&86.4&78.9&83.6&87.6\\ 
                Fog& 91.8&60.3&78.6&86.9\\ 
                Gaussian Blur& 91.4&78.0&85.4&89.7\\ 
                Gaussian Noise&78.6&79.1&85.9&90.4\\ 
                Glass Blur&71.7&77.9&80.9&82.6\\ 
                Implus Noise&76.1&73.8&81.9&86.0\\ 
                JPEG Compression&78.8&82.8&85.8&89.9\\ 
                Motion Blur & 89.6&76.5&84.1&88.3\\ 
                Pixelate& 88.3&82.7&86.1&90.1\\
                Saturate& 93.3&81.5&87.3&91.4\\ 
                Shot Noise & 81.9&80.4&86.2&90.8\\ 
                Snow& 86.3&80.4&84.0&89.0\\ 
                Spatter& 88.3&80.7&84.1&87.8\\ 
                Speckle Noise& 82.1&80.2&86.0&90.6\\ 
                Zoom Blur& 91.1&78.9&86.0&90.2\\ 
              \hline
                Mean &86.3 &76.6&84.6&\textbf{89.0}\Tstrut\\ 
                Variance &41.2 &90.2&5.5&\textbf{4.9}\\ 
			\bottomrule
		\end{tabular}}
		\vspace{0.2em}
	\caption{Comparative analysis of robustness (\%) to common corruptions is shown. SAT shows significant improvement over majority of the corruptions and does specially well against those that are most difficult for naturally trained (NT) models such as glass blur, Gaussian noise and impulse noise. Mean accuracy (higher is better) and variance (lower is better) are reported.}
    \label{tab:results_on_common_corruptions}
\end{table}

\subsubsection{Robustness to Common Corruptions} Distributional shifts in the data can also come in the form of natural corruptions \cite{hendrycks2019robustness}. Hendrycks \etal \cite{hendrycks2019robustness} simulated multiple such corruptions including snow, fog and glass blur. We study 18 of such corruptions. Depending upon the severity, each corruption is sub-divided into 5 levels resulting in a total of 50k images for every corruption type. Analysis of robust models on such distributional shifts is presented in Table \ref{tab:results_on_common_corruptions}. Interestingly, theoretically robust model, Trades \cite{Zhang2019theoretically} loses significant accuracy on such corruptions as compared to naturally trained (NT) models. Feature scattering performed better than Trades. We observe that, compared with Trades \cite{Zhang2019theoretically} and FS \cite{feature_scatter}, a naturally trained model shows better robustness to these image corruptions. Our proposed SAT, however, demonstrates improved generalization to common corruptions while simultaneously providing adversarial robustness. Contrastive nature of our adversarial training helps the model to adapt to such distribution shifts which aligns with findings that discriminative learning boosts domain adaptation and generalization \cite{motiian2017unified}. Further, it is consistent with the experiments where random sterilizations also help in boosting domain adaptation \cite{jackson2019style}. We further compare with \cite{xie2020adversarial} that uses adversarial examples to regularize the model to increase generalization to common corruptions. Xie \etal \cite{xie2020adversarial} use a small perturbation budget e.g., $\epsilon \le 4$ to avoid drop in clean accuracy, while our method (SAT) is capable of handling larger distortions ($\epsilon=8$ for SVHN, CIFAR10 and CIFAR100) while introducing minimal degradation to clean accuracy.  We train on WideResNet for CIFAR10 using recommended settings from \cite{xie2020adversarial}. It provides 90.2\% mean corruption accuracy which is only $\sim$1\%  better than SAT however \cite{xie2020adversarial} provides  
0\% adversarial robustness in comparison to SAT ($n=1$) that offers 13.4\% against Auto-Attack at $\epsilon=8$. We further observe that in comparison to the recent method such as LBGAT \cite{cui2020learnable} that performs better on complex data distributions using distillation from the teacher model, our method's performance improves with data augmentation \cite{rebuffi2021fixing} (Table \ref{tab:new_results_using_data_augmentation}).

\begin{figure}[t]
\centering
  \begin{minipage}{.24\textwidth}
  	\centering
    \includegraphics[  width=\linewidth, keepaspectratio]{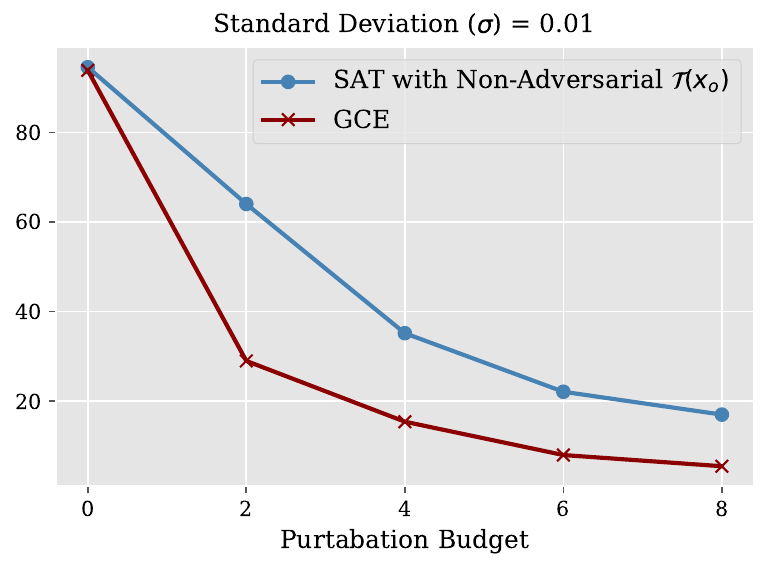}
  \end{minipage} 
  \begin{minipage}{.24\textwidth}
  	\centering
    \includegraphics[width=\linewidth, keepaspectratio]{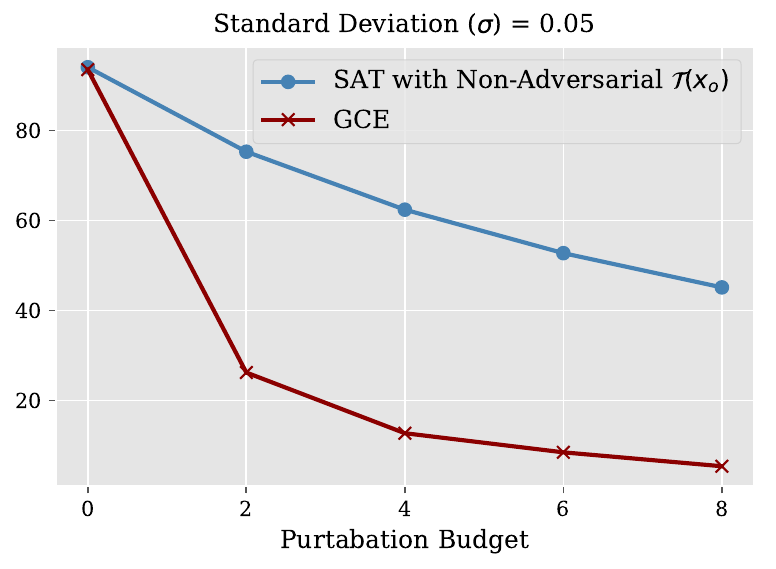}
  \end{minipage} 
  \caption{\textbf{White-box analysis:} Our approach with non-adversarial transformation is compared against GCE \cite{chen2019improving}. Robustness is measured on CIFAR-10 test set against PGD with 20 iterations and random restarts. Models trained with our approach are significantly robust compared to GCE method \cite{chen2019improving}.}
\label{fig:cat_vs_gce_white_box}
\end{figure}

\begin{table}[t]
	\centering\setlength{\tabcolsep}{2pt}
			\scalebox{0.85}{
			\begin{tabular}{lcccccc}
				\toprule
				Method & Model & Parameters (M)  & Training & CutMix & Clean & AA \\
				\midrule
				\multicolumn{7}{c}{Dataset $\rightarrow$ CIFAR10} \\
				\midrule
		Gowal et al. \cite{rebuffi2021fixing} & WRN & 36 & Trades \cite{Zhang2019theoretically} & \cmark & 86.09 & 57.50\\
		SAT ($n=10$) &  WRN & 36 & Stylized Attack &\cmark & \textbf{90.80} & \textbf{58.61}\\
		\midrule
		\multicolumn{7}{c}{Dataset $\rightarrow$ Tiny-ImageNet} \\
		\midrule
		Gowal et al. \cite{rebuffi2021fixing} & WRN & 36 & Trades \cite{Zhang2019theoretically} & \cmark & 53.69 & 23.83\\
		SAT ($n=10$) &  WRN & 36 & Stylized Attack &\cmark & \textbf{55.86} & \textbf{25.04}\\
		
			\bottomrule
		\end{tabular}}
		\vspace{0.2em}
	\caption{The use of data augmentation improves the robustness of our SAT (WRN-28-10 \cite{rebuffi2021fixing}) by a large margin. We follow the default settings as proposed by \cite{rebuffi2021fixing} for applying CutMix.}
    \label{tab:new_results_using_data_augmentation}
    \vspace{-1em}
\end{table}

\subsection{Effect of Data Augmentations}
Rebuffi et.al \cite{rebuffi2021fixing, rebuffi2021data} observed the increase in adversarial robustness by carefully tuning the data augmentations during training. Our approach scales well with the data augmentation proposed in \cite{rebuffi2021fixing} and increases the adversarial robustness of our stylized adversarial training (Table \ref{tab:new_results_using_data_augmentation}).

\subsection{Non-adversarial $\mathcal{T}$: Defense Results and Insights}\label{subsec:results_abstract_training_framework}
\begin{figure}[t]
\centering
  \begin{minipage}{.23\textwidth}
  	\centering
    \includegraphics[  width=\linewidth, keepaspectratio]{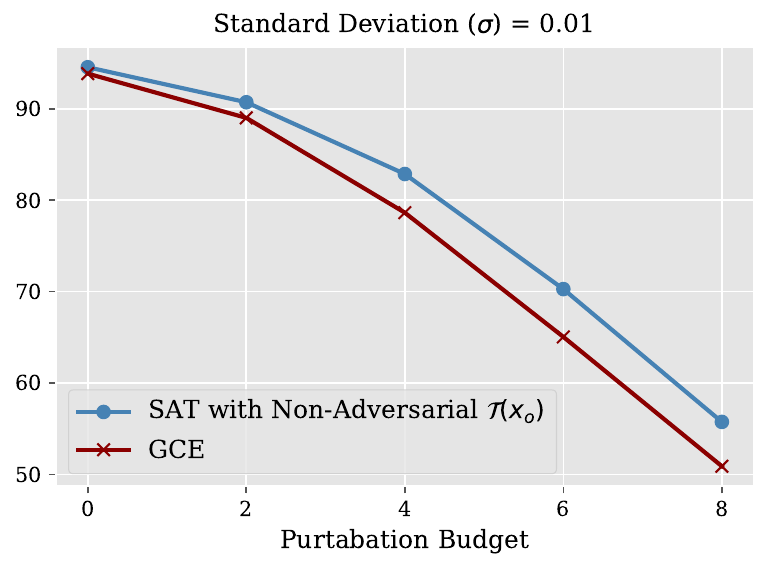}
  \end{minipage} 
  \begin{minipage}{.23\textwidth}
  	\centering
    \includegraphics[width=\linewidth, keepaspectratio]{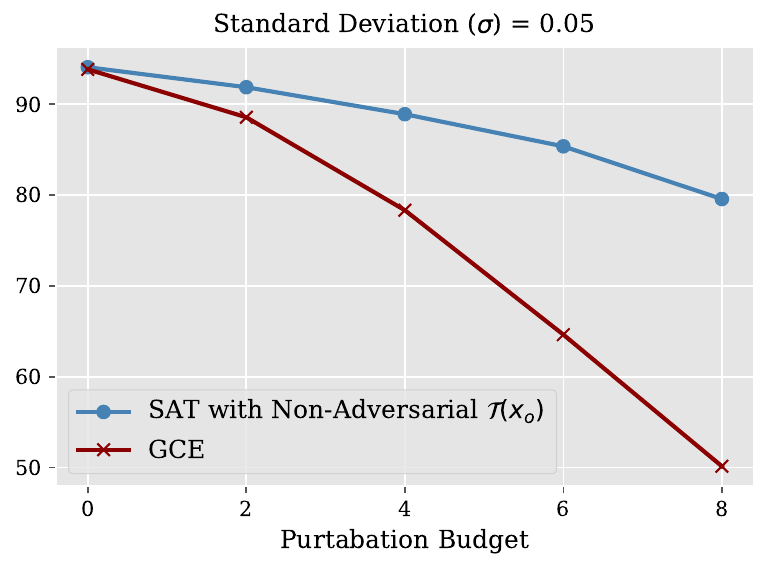}
  \end{minipage} 
  \caption{\textbf{Black-Box analysis:} Our approach trained with non-adversarial perturbations is compared against GCE \cite{chen2019improving}. Adversaries are generated using MIFGSM with 10 iterations on CIFAR-10 test set. Our trained models show high resistance to transferable attack as compared to GCE \cite{chen2019improving}.}
\label{fig:cat_vs_gce_black_box}
\end{figure}

In this section, we analyse the performance of training SAT with non-adversarial transformation to establish empirical evidence of adversarial robustness with non-adversarial transformation. For this case, Gaussian noise is considered  and ResNet18 \cite{he2016deep} is trained using our approach and GCE \cite{chen2019improving} on CIFAR-10 training set. White-box robustness is measured against PGD \cite{madry2018towards} with 20 iterations. Black-box robustness is evaluated by transferring adversaries from VGG19 \cite{Simonyan14verydeep} using momentum iterative fast sign gradient (MIFSGM) \cite{dong2018boosting} attack on the CIFAR-10 test set. From our experimenatl results shown in Figures \ref{fig:cat_vs_gce_white_box} and \ref{fig:cat_vs_gce_black_box}, we observe that, as the transformation, $\mathcal{T}$, becomes better \eg by increasing the standard deviation ($\sigma$) of Gaussian noise, our robustness against PGD attack increases significantly. This suggests that better transformation can lead to more robust models.
We further notice that the Gaussian transformation does not have noticeable effect on GCE performance in terms of robustness.
It is interesting to note that our approach maintains accuracy on clean examples while its robustness improves with better transformation. This behavior complements our design approach that takes into account the relationship of original and transformed samples unlike GCE \cite{li2019improving} which only relies on minimizing the probability of other classes with respect to the true class.

\begin{figure}[t]
\centering
  	\centering 
    \includegraphics[width=0.8\linewidth, keepaspectratio]{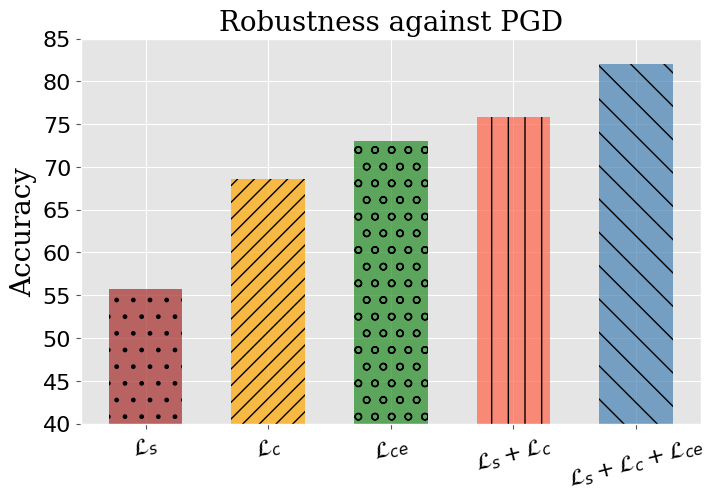}
    \vspace{-0.5em}
  \caption{White-box robustness analysis shows the effectiveness of different losses introduced in  Eq. \ref{eq:adv_generation_loss}. Results are reported for WideResNet on CIFAR10 dataset for single-step SAT (Algorithm \ref{alg:rtf}).}
\label{fig:loss_abaltion}
\end{figure}

\begin{figure*}[t]
\centering
\begin{minipage}{0.19\textwidth}
  	\centering
    \includegraphics[height=2.5cm, width=\linewidth]{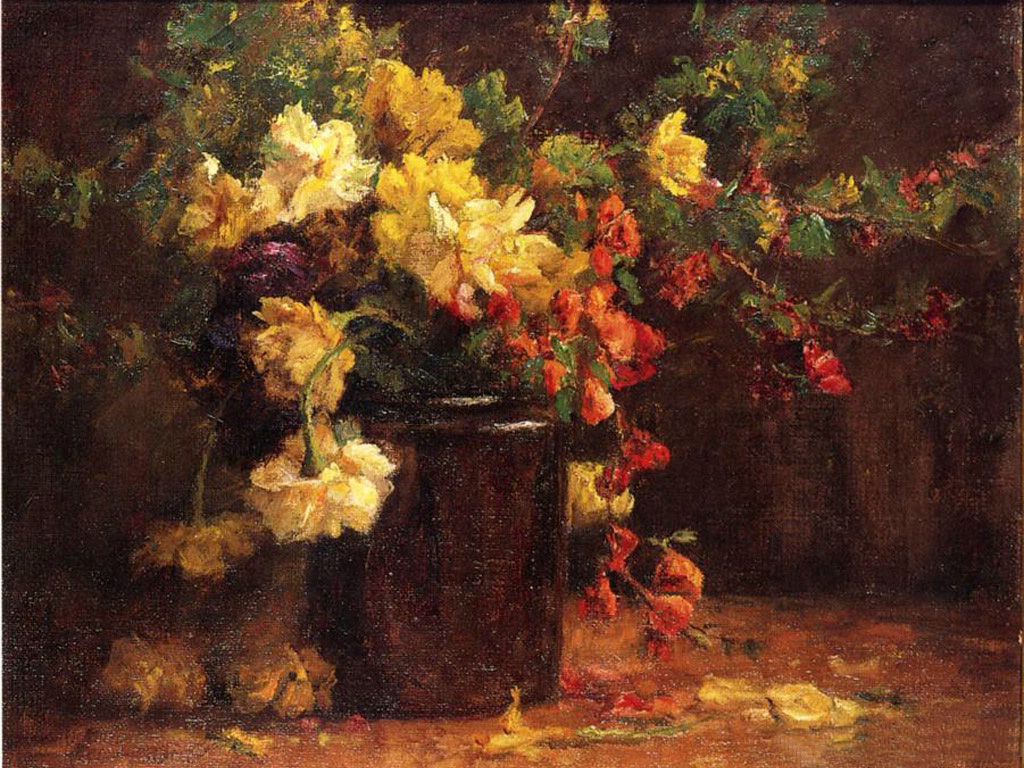}
  \end{minipage} 
  \begin{minipage}{0.19\textwidth}
  	\centering
    \includegraphics[height=2.5cm, width=\linewidth]{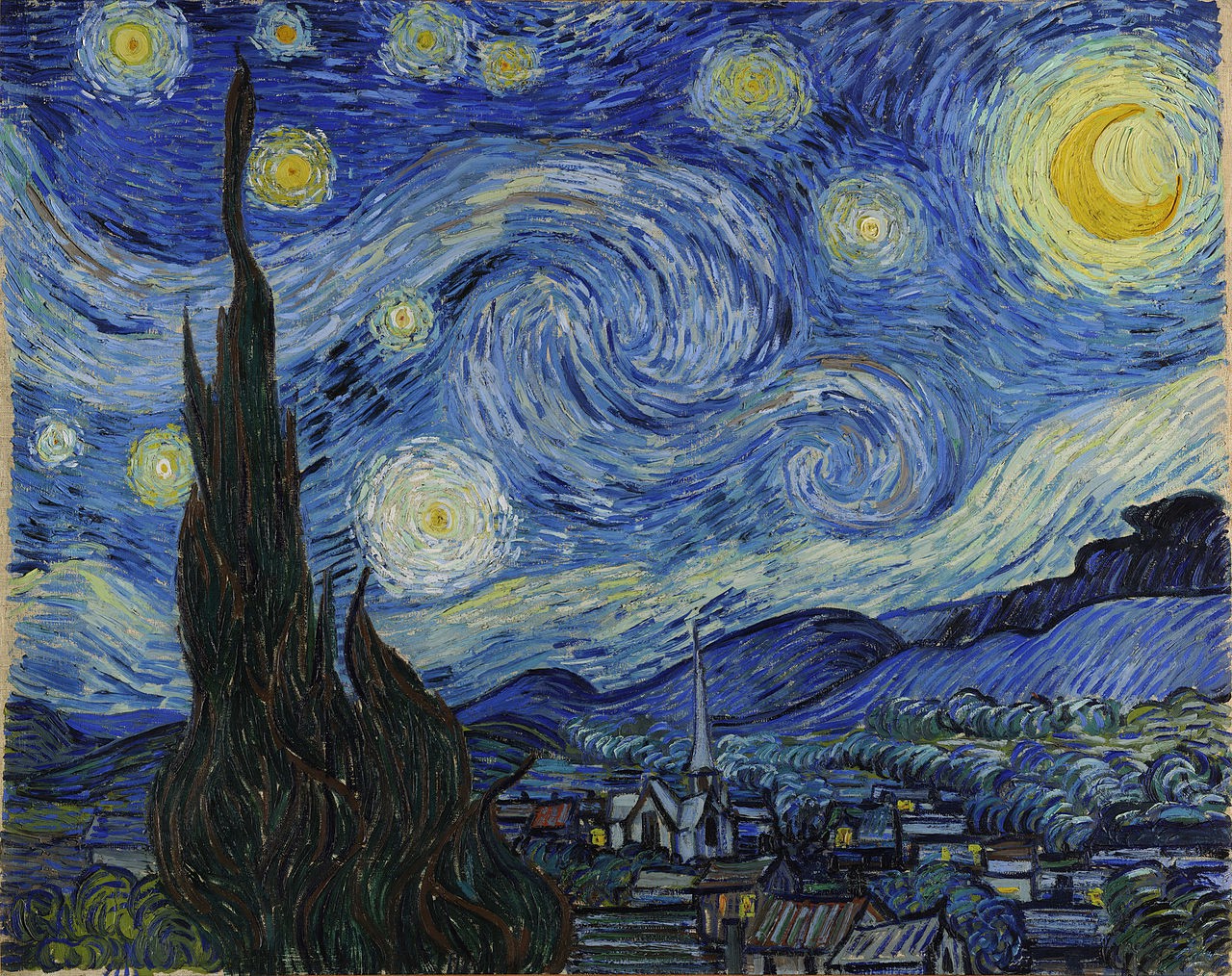}
  \end{minipage} 
  \begin{minipage}{0.19\textwidth}
  	\centering
    \includegraphics[height=2.5cm, width=\linewidth]{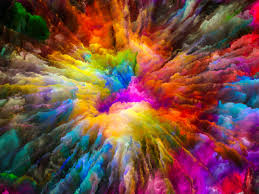}
  \end{minipage} 
    \begin{minipage}{0.19\textwidth}
  	\centering
    \includegraphics[height=2.5cm, width=\linewidth,trim=0mm 2mm 0mm -2mm]{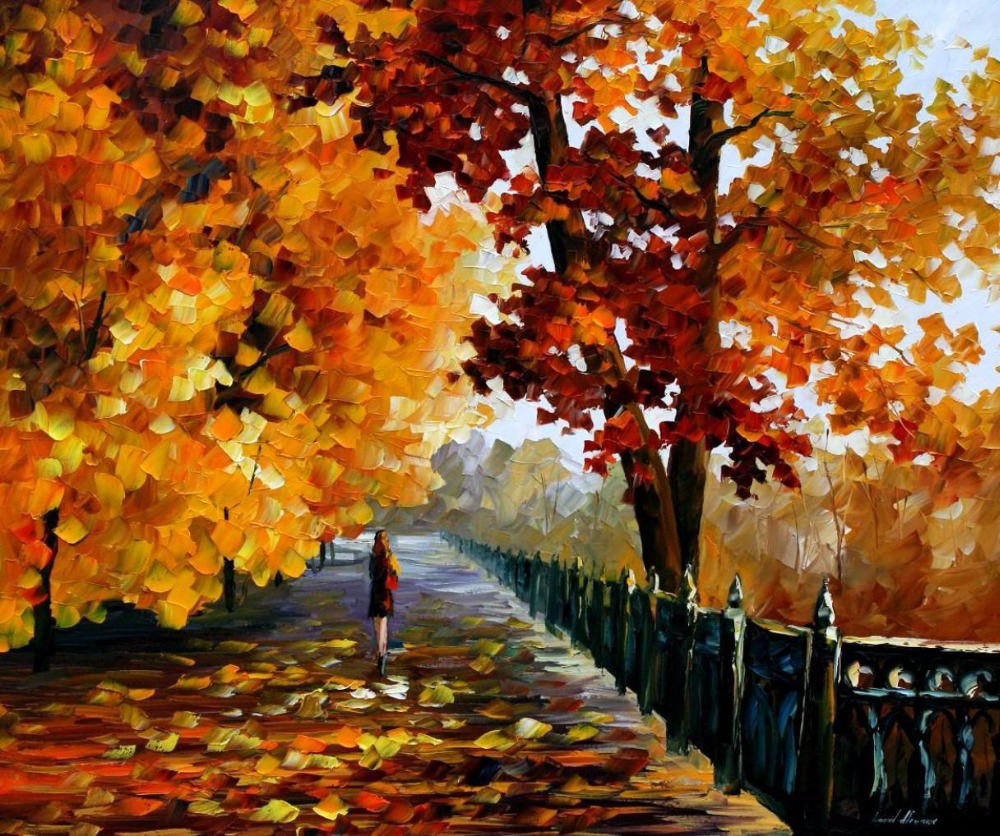}
  \end{minipage} 
  \begin{minipage}{0.19\textwidth}
  	\centering
    \includegraphics[ height=2.5cm, width=\linewidth, trim=0mm 2mm 0mm -2mm]{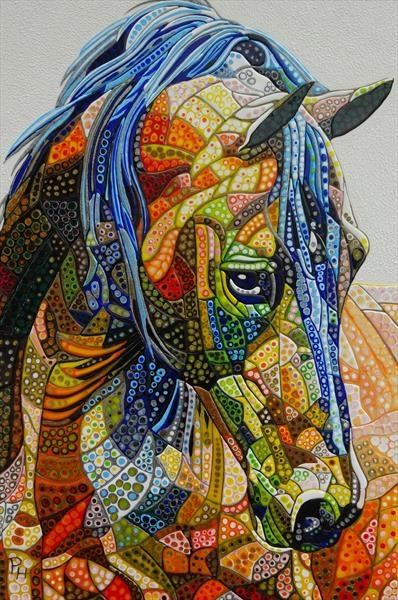}
  \end{minipage} 
  \\\vspace{0.5ex}
  \begin{minipage}{0.19\textwidth}
  	\centering
    \includegraphics[height=2.5cm, width=\linewidth]{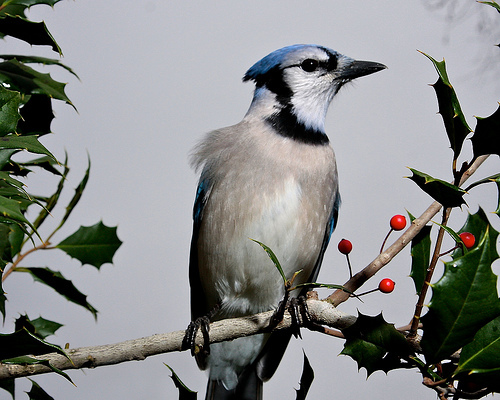}
  \end{minipage} 
  \begin{minipage}{0.19\textwidth}
  	\centering
    \includegraphics[height=2.5cm, width=\linewidth]{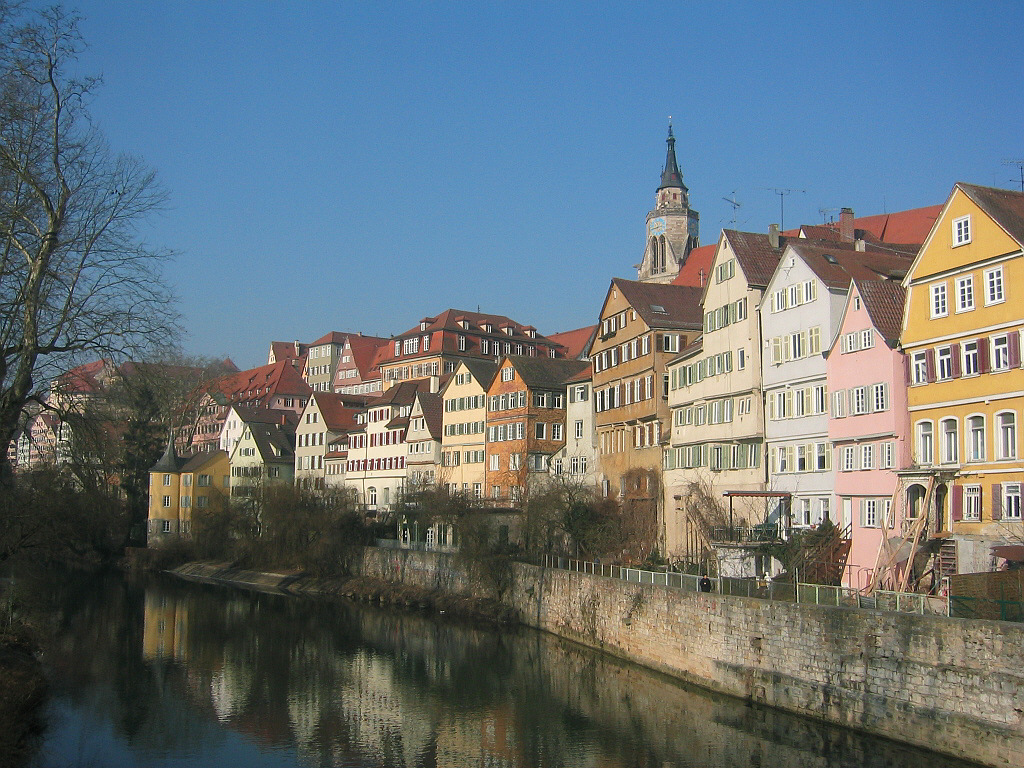}
  \end{minipage} 
  \begin{minipage}{0.19\textwidth}
  	\centering
    \includegraphics[height=2.5cm, width=\linewidth]{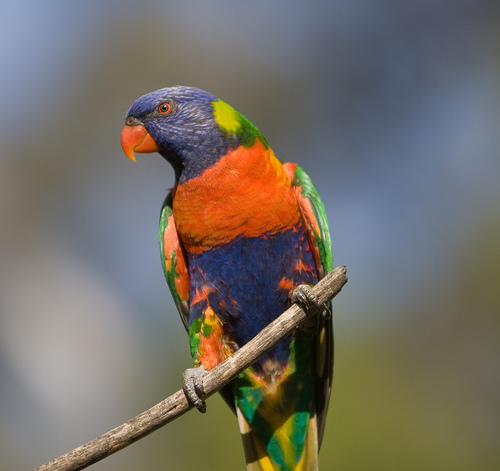}
  \end{minipage} 
    \begin{minipage}{0.19\textwidth}
  	\centering
    \includegraphics[height=2.5cm, width=\linewidth]{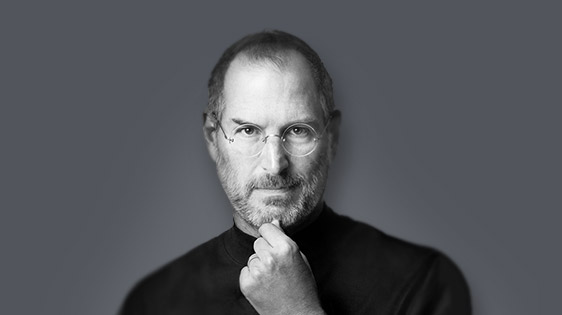}
  \end{minipage} 
   \begin{minipage}{0.19\textwidth}
  	\centering
    \includegraphics[ height=2.5cm, width=\linewidth]{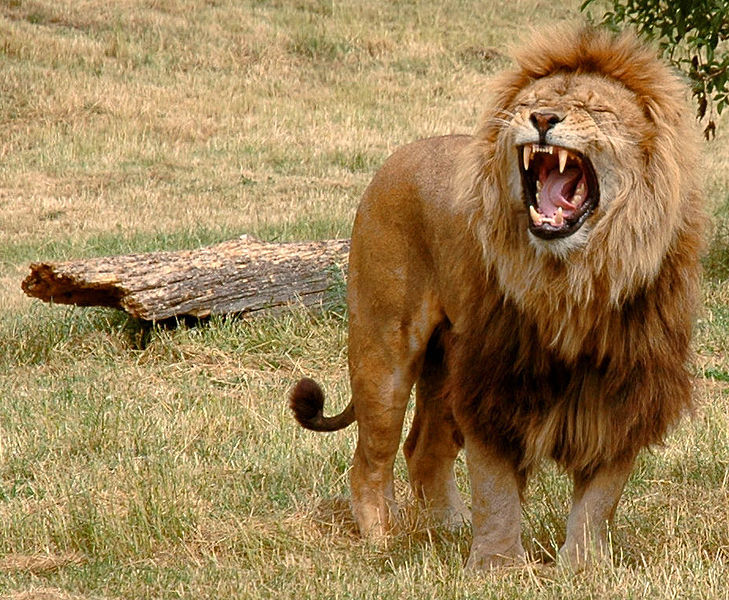}
  \end{minipage} 
  \\\vspace{0.5ex}
  \begin{minipage}{0.19\textwidth}
  	\centering
    \includegraphics[height=2.5cm, width=\linewidth]{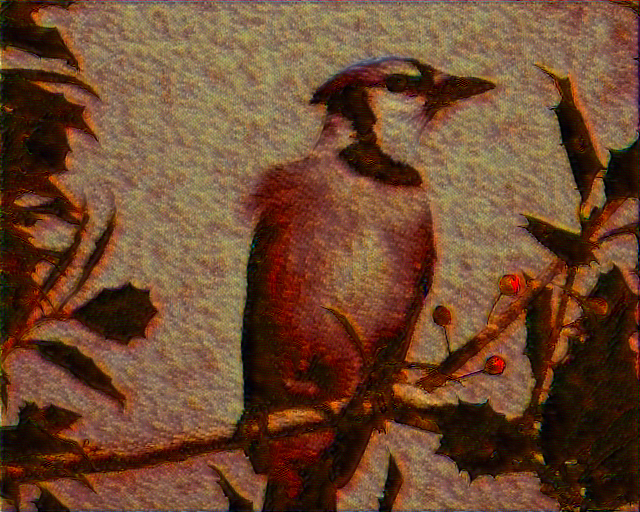}
  \end{minipage} 
  \begin{minipage}{0.19\textwidth}
  	\centering
    \includegraphics[height=2.5cm, width=\linewidth]{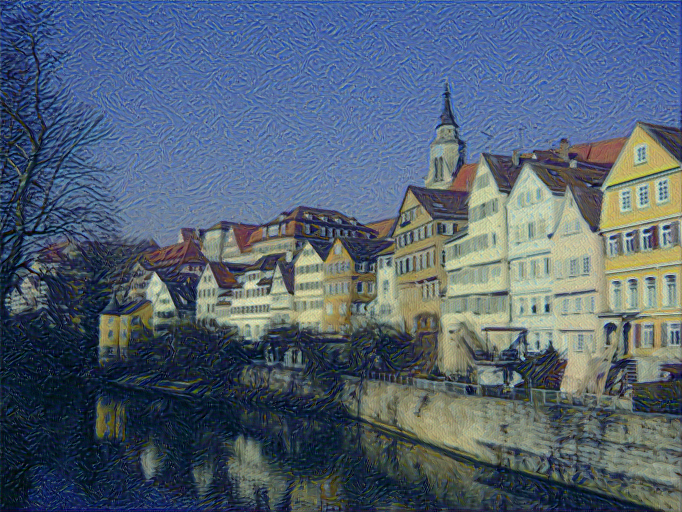}
  \end{minipage} 
  \begin{minipage}{0.19\textwidth}
  	\centering
    \includegraphics[height=2.5cm, width=\linewidth]{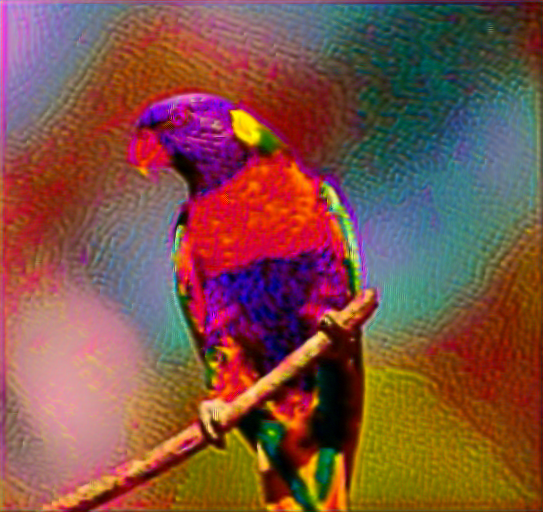}
  \end{minipage} 
    \begin{minipage}{0.19\textwidth}
  	\centering
    \includegraphics[height=2.5cm, width=\linewidth]{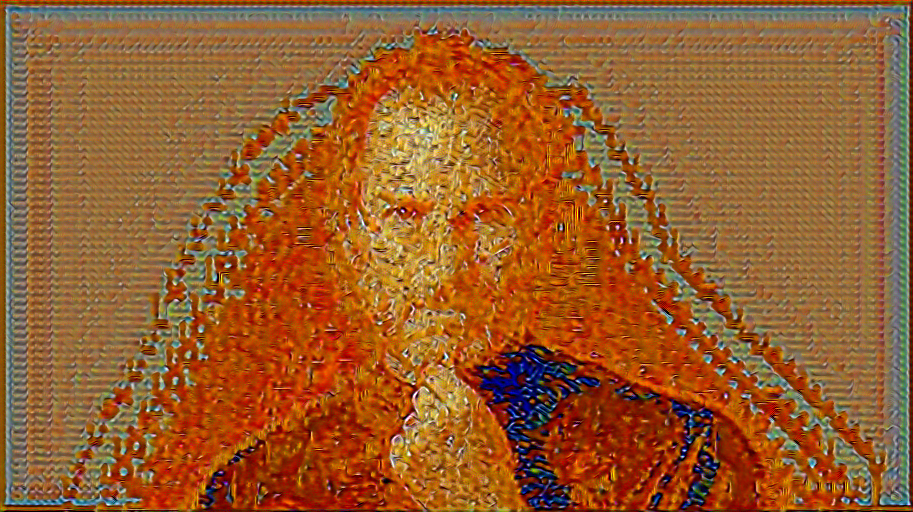}
  \end{minipage} 
   \begin{minipage}{0.19\textwidth}
  	\centering
    \includegraphics[ height=2.5cm, width=\linewidth]{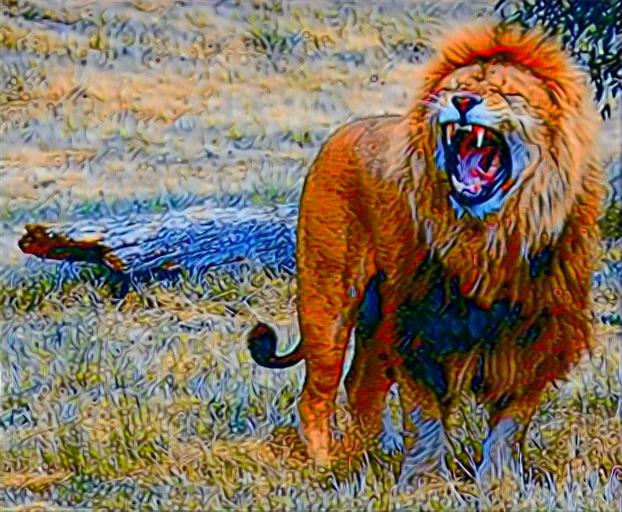}
  \end{minipage} 
   \\\vspace{0.5ex}
   \begin{minipage}{0.19\textwidth}
  	\centering
    \includegraphics[height=2.5cm, width=\linewidth]{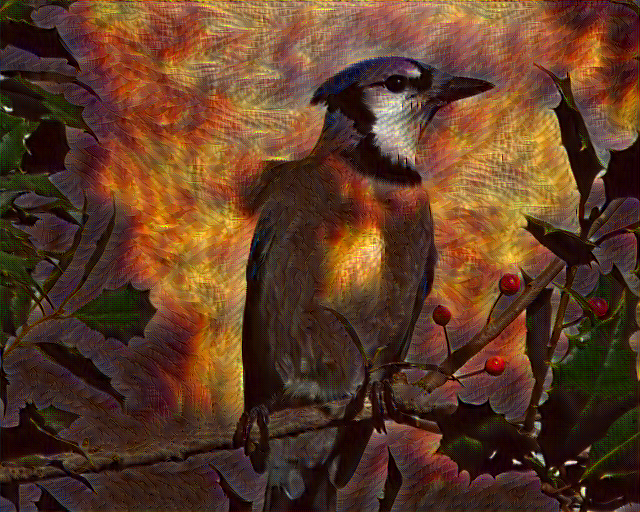}
  \end{minipage} 
  \begin{minipage}{0.19\textwidth}
  	\centering
    \includegraphics[height=2.5cm, width=\linewidth]{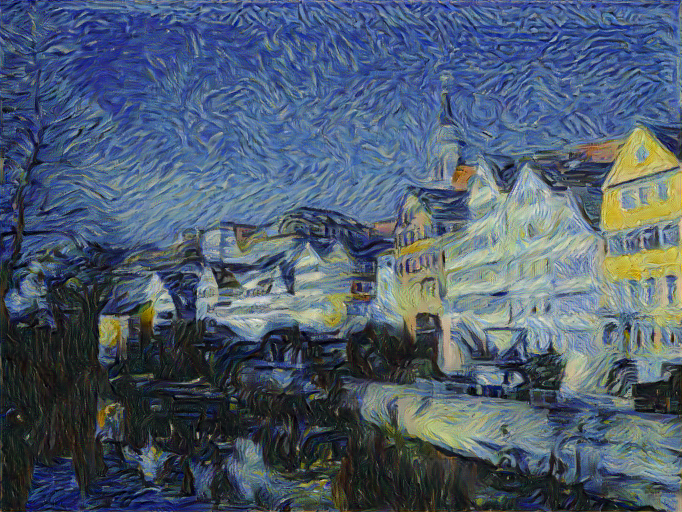}
  \end{minipage} 
  \begin{minipage}{0.19\textwidth}
  	\centering
    \includegraphics[height=2.5cm, width=\linewidth]{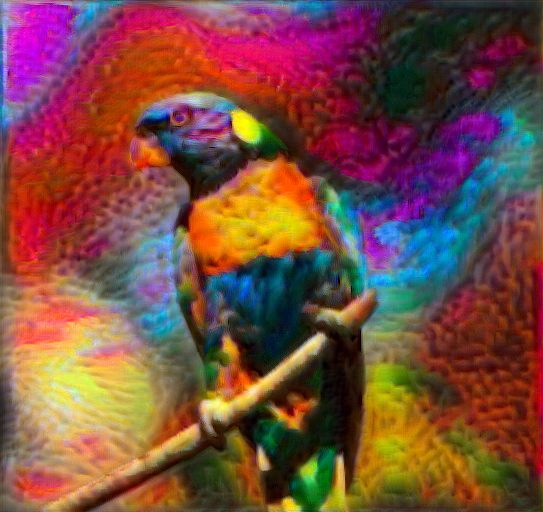}
  \end{minipage} 
    \begin{minipage}{0.19\textwidth}
  	\centering
    \includegraphics[height=2.5cm, width=\linewidth]{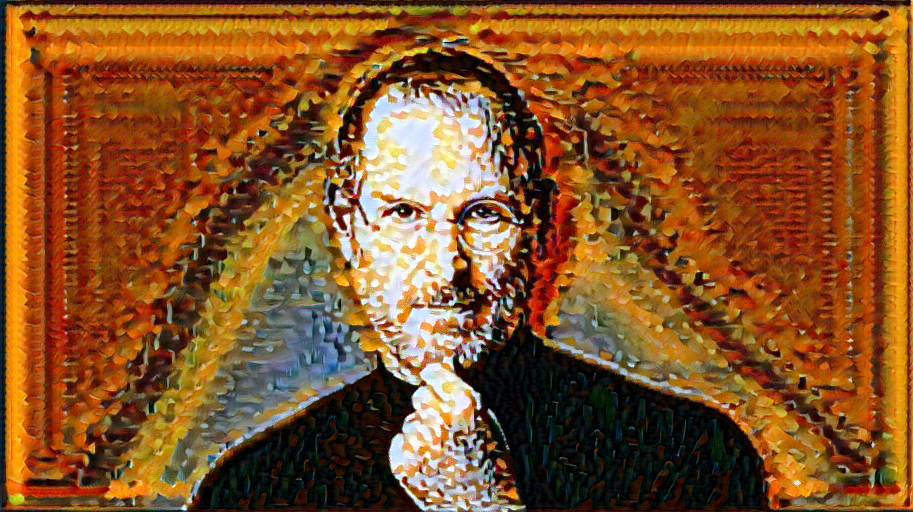}
  \end{minipage} 
   \begin{minipage}{0.19\textwidth}
  	\centering
    \includegraphics[ height=2.5cm, width=\linewidth]{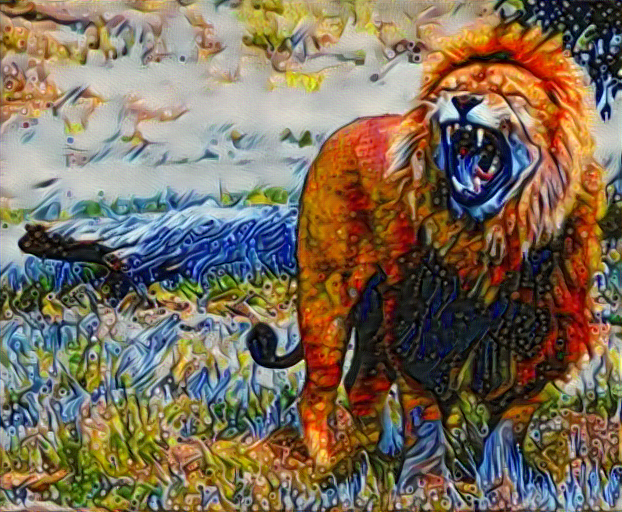}
  \end{minipage} 
  \caption{ \textbf{Style transfer} using features of ResNet18 trained on CIFAR10 dataset. Robust features obtained using SAT produced more perceptually appealing style transfer as compared to naturally trained (NT) feature space of ResNet18. Comparison is made under the same hyper-parameters and number of iterations (100). \emph{(Top to bottom)} 1st and 2nd rows show style and target images while 3rd and 4th rows show style transfer using NT and SAT models, respectively. }
  \label{fig:style_transfer}
\end{figure*}

\subsection{Ablation Study}
\label{sec:ablations}
We dissect our proposed adversarial training for WideResNet architecture on CIFAR10 dataset to develop further insights, as follows:

\textbf{Contribution of proposed losses:} We show in Figure \ref{fig:loss_abaltion}  that each loss proposed in Eq. \ref{eq:adv_generation_loss} contributes towards SAT's robustness. Individually, the content loss demonstrates better robustness compared to style loss while the classification loss (CE) provides better robustness than content loss. One potential reason for this behavior is that the style loss encodes more abstract information about the target sample compared to content and classification losses. Interestingly, the combination of style and content losses could beat the case when only CE loss is used. Overall, SAT performance increases significantly when adversaries are computed using style, content and boundary information of the target samples.  

\textbf{Hyper-parameters:} Hyper-parameter search is conducted on CIFAR10 and these parameters are kept the same for CIFAR100 and SVHN datasets. We created a validation set of 5k samples from CIFAR10 training set. SAT is trained on rest of the 45k images by varying the weightage of $\alpha$, $\gamma$, $\beta$ (Eq.~\ref{eq:adv_generation_loss}). In general, SAT achieved best results (Table ~\ref{tab:hyper_paramters}) with equal weightage, i.e., $\alpha = \gamma = \beta$. All hyper-parameters are tuned with computationally efficient version of our approach, SAT ($n=1$), on 5k validation samples.

\begin{table}[t]
	\centering\setlength{\tabcolsep}{5pt}
		\resizebox{0.75\linewidth}{!}{
		\begin{tabular}{ l|c|c|c|c}
		\toprule
		& \rotatebox[origin=c]{90}{$\alpha=0.5$}\hspace{1em} \rotatebox[origin=c]{90}{$\gamma=0.5$} & \rotatebox[origin=c]{90}{$\alpha=1.0$} \hspace{1em}\rotatebox[origin=c]{90}{$\gamma=0.5$} & \rotatebox[origin=c]{90}{$\alpha=0.5$}\hspace{1em} \rotatebox[origin=c]{90}{$\gamma=1.0$}&\rotatebox[origin=c]{90}{$\alpha=1.0$} \hspace{1em}\rotatebox[origin=c]{90}{$\gamma=1.0$}\\
		\midrule
	    $\beta=0.5$&79.1&71.1&74.9&77.3\\
	  $\beta=1.0$&76.0&76.8&78.0&82.5 \\
		\bottomrule
		\end{tabular}
	}
	\vspace{0.2em}
	\caption{\textbf{Hyper-Parameter} search. Single-step SAT is trained on CIFAR10 and evaluated on validation set (5k samples selected from the training set) against PGD attack with 10-iterations.}
	\label{tab:hyper_paramters}
\end{table}
 
 \begin{table}[t]
  \centering\setlength{\tabcolsep}{5pt}
  \scalebox{1.1}{
  \begin{tabular}{ccccc}
		\toprule
		Number of Epochs ($\rightarrow$)&100&  150 & 200 & 250\\
		\midrule
		Clean ($\rightarrow$)&91.0&92.5&93.3&93.3\\
		PGD ($\rightarrow$)&77.5&	80.0	&81.1	&81.1\\
		Corruptions ($\rightarrow$)&85.7 & 87.2 & 89.0 & 89.0\\
		\bottomrule
\end{tabular}}
\vspace{0.2em}
\caption{ Convergence analysis of single-step SAT.}
\label{tab:convergence}
\vspace{-2em}
\end{table}

\textbf{Convergence Analysis:} Each SAT model is trained for 200 epochs. We set the initial learning rate to 0.1 and decrease it by a factor of 0.1 at epochs 50 and 95. We observe that robustness increases with the number of iterations as shown in Table \ref{tab:convergence}. We did not observe a noticeable gain in training beyond 200 epochs.

\begin{table}[t]
  \centering\setlength{\tabcolsep}{3pt}
  \scalebox{0.95}{
  \begin{tabular}{ccccc}
		\toprule
		SAT ($n=1$) & SAT ($n=10$) & Madry \etal \cite{madry2018towards}&  Trades \cite{Zhang2019theoretically} & FS\cite{feature_scatter}\\
		\midrule
		11.6 & 32.2 & 23.6 & 29.2&29.4\\
		\bottomrule
\end{tabular}}
\vspace{0.2em}
\caption{ Training time (hours) on a Tesla V100.}
\label{tab:train_time_comparision}
\vspace{-2em}
\end{table}

\textbf{Computational Training Cost:} As mentioned above, SAT with $n=1$ takes 200 epochs for convergence which takes around 11.6 hours of training time (Table \ref{tab:train_time_comparision} for computational analysis). This cost is less than the adversarial training approach of \cite{madry2018towards}, Trades and FS. This is because SAT with $n=1$ requires only one attack iteration to compute adversaries. Multi-step SAT is relatively costly as it depends on 10 stylized attack iterations (Table \ref{tab:train_time_comparision}).

\textbf{Random Priors:} We study the effect of target attacks in PCL \cite{Mustafa_2019_ICCV} and untarget attack on the performance of SAT training framework. We observe that adversarial robustness of PCL and SAT decreases when combined with target and untarget attacks, respectively (Table ~\ref{tab:SAT_with_untarget}). We hypothesis the reason for the decreased robustness is the random priors when we switch the attack type in PCL and SAT training framework. If PCL uses targeted attack, then target attack label for each sample will be randomly selected. For example, for each sample in CIFAR10 training set, there are 9 target attack labels available for each source class. These random targeted adversaries might not be optimal for PCL as PCL framework does not exploit the target label information during training and only relies on original labels of natural samples. In contrast, our framework (SAT) sets the target samples as a prior, so if SAT uses untarget attack then our prior (target samples) will be random which is not optimal for SAT. Furthermore, such untarget adversaries are based on model's boundary information and do not contain style or content information. SAT leads to best results when it exploits target samples used to create adversaries during training. 

\begin{table}[t]
	\centering\setlength{\tabcolsep}{5pt}
		\resizebox{0.75\linewidth}{!}{
		\begin{tabular}{ l c c}
		\toprule
		Method & Clean & PGD\\
		\midrule
	   PCL \cite{Mustafa_2019_ICCV} with Targeted Attack & 92.0 & 38.7\\
	 SAT ($n=1$) with Untargeted Attack & 90.3 & 66.4\\
		\bottomrule
		\end{tabular}
	}
	\vspace{0.2em}
	\caption{\textbf{Random Priors}. SAT is trained with an untargeted attack, so target samples used in SAT become a random prior in relation to the adversarial noise. PCL is trained on targeted attacks and targeted attack labels are chosen randomly during training. This study is conducted on CIFAR10. PGD ran for 10 iterations. We observed that robustness of both SAT and PCL decreases.}
	\label{tab:SAT_with_untarget}
\end{table}

\begin{table}[t]
      \centering\setlength{\tabcolsep}{7pt}
      \scalebox{1.1}{
  	\begin{tabular}{ l|cc|cc|cc}
			\toprule
			\multirow{2}{*}{Source}& \multicolumn{2}{c}{MIFGSM}&  \multicolumn{2}{c}{PGD} & \multicolumn{2}{c}{CW}\\
			\cline{2-7} 
			&  FS \cite{feature_scatter} & SAT &  FS\cite{feature_scatter} & SAT & FS\cite{feature_scatter} & SAT \Tstrut\\
			\hline
			NT & 87.5&\textbf{88.3}&89.0&\textbf{91.5}&88.1&\textbf{90.3}\\
			AT & 80.8&\textbf{81.6}&80.0&\textbf{83.5}&79.5&\textbf{82.4} \\ 
			\bottomrule
  \end{tabular}}
  \vspace{0.2em}
  \caption{\textbf{Black-box} robustness evaluation. Adversaries are transferred from naturally and adversarially \cite{feature_scatter} trained models with the same architectures as of SAT and FS \cite{feature_scatter} (WideResNet). High accuracy on black-box adversaries indicate model convergence on non-degenerate solution that is without gradient masking.}
  \label{tab:rtf_black_box_acc}
\end{table}

\subsection{Sanity Checks for Gradient Obfuscation}
Gradient obfuscation or gradient masking refers to the phenomenon where optimization based attacks fail thus leading to high but false adversarial robustness. Athalye \etal \cite{obfuscated-gradients} devise certain tests to evaluate if the defense is relying on gradient masking. We perform the following sanity checks on our defenses.
\begin{itemize}
    \item \textit{\textbf{Robustness to Black-Box Attacks:}} If black-box attack (where adversaries are transferred from another model) are stronger than the white-box, this indicates that white-box attack is weak and gradients are being obfuscated. We evaluated SAT (Algorithm \ref{alg:rtf}) under different black-box attacks (Table \ref{tab:rtf_black_box_acc}) and the accuracy of our defense remains higher than the white-box attacks (Table \ref{tab:cifar10_comparision_with_fs}).
    \item \textit{\textbf{Iterative attack should be stronger than single-step:}} Another test for gradient masking is that iterative attacks like PGD with small step-size should be more effective than single step attack like FGSM. In all of our evaluations (Tables \ref{tab:comparison_with_trades}, \ref{tab:cifar10_comparision_with_fs}, \ref{tab:cifar100_comparision_with_fs} and \ref{tab:comparison_with_pcl}), PGD with step size 2/255 is always a stronger attack than FGSM. 
    \item \textit{\textbf{Robustness should approach to zero for large enough perturbation:}} Gradient masking occurs if defense accuracy does not approach to zero for large enough perturbation. Our defense also fulfills this sanity check as their accuracy goes to zero at $\epsilon=128$.
    \item \textit{\textbf{Evaluation with Ensemble of Attacks :}} Performance gap between a simple PGD attack and an ensemble of attacks (Auto-Attack) is only marginal for our SAT with $n=10$ (Tables \ref{tab:comparison_with_trades}, \ref{tab:cifar100_comparision_with_fs}, and \ref{tab:comparison_with_pcl}). This behavior is consistent with other masking free methods (PAT \cite{madry2018towards} and Trades \cite{Zhang2019theoretically}). Therefore, our SAT does not suffer from gradient obfuscation as it shows consistent robustness regardless of the attack type.
\end{itemize}

\subsection{Improved Style Transfer with SAT}

Image style transfer works well for VGG features compared with residual connection based models \cite{nakano2019a}. For the case of residual networks, it has been noted that compared with their naturally trained counterpart, features from adversarially robust models generate more visually appealing images for style transfer \cite{nakano2019a}. In our case, we observe (Figure \ref{fig:style_transfer}) that ResNet18 trained on CIFAR10 using SAT performs a better style transfer, compared with a naturally trained ResNet18. This indicates that our approach learns representations that can faithfully model the perceptual space.

\section{Conclusion}
We propose to maximize inter-class margins by setting target class samples as priors to adversarial perturbations for the original samples. Our framework pushes the clean images towards randomly selected targets by adding the style, content and boundary information of the target image from other classes in the form of adversarial perturbations within an allowed perturbation budget. Our framework naturally fits within max-margin learning as it generates positive (adversaries) and negatives (target samples) for clean images. Adversarially trained models using our framework show significant robustness against adversarial attacks (both in white-box and black-box attack scenarios), naturally occurring distributional shifts as well as on common corruptions. Furthermore, robust features obtained via our proposed approach can also be used for style transfer. Our approach successfully demonstrates that stylized targeted attack can boost adversarial robustness  with minimal negative impact on generalization . In future,  we will study the impact of popular heuristics used in the litterateur \eg using synthetic data \cite{gowal2021improving}, pertaining \cite{hendrycks2019using}, distillation from naturally trained model \cite{cui2020learnable} and search over suitable activation functions \cite{gowal2020uncovering} which can potentially further improve our model's robustness.

\bibliographystyle{IEEEtran}
\bibliography{egbib}

\begin{thebibliography}{10}
\providecommand{\url}[1]{#1}
\csname url@samestyle\endcsname
\providecommand{\newblock}{\relax}
\providecommand{\bibinfo}[2]{#2}
\providecommand{\BIBentrySTDinterwordspacing}{\spaceskip=0pt\relax}
\providecommand{\BIBentryALTinterwordstretchfactor}{4}
\providecommand{\BIBentryALTinterwordspacing}{\spaceskip=\fontdimen2\font plus
\BIBentryALTinterwordstretchfactor\fontdimen3\font minus
  \fontdimen4\font\relax}
\providecommand{\BIBforeignlanguage}[2]{{%
\expandafter\ifx\csname l@#1\endcsname\relax
\typeout{** WARNING: IEEEtran.bst: No hyphenation pattern has been}%
\typeout{** loaded for the language `#1'. Using the pattern for}%
\typeout{** the default language instead.}%
\else
\language=\csname l@#1\endcsname
\fi
#2}}
\providecommand{\BIBdecl}{\relax}
\BIBdecl

\bibitem{madry2018towards}
\BIBentryALTinterwordspacing
A.~Madry, A.~Makelov, L.~Schmidt, D.~Tsipras, and A.~Vladu, ``Towards deep
  learning models resistant to adversarial attacks,'' in \emph{International
  Conference on Learning Representations}, 2018. [Online]. Available:
  \url{https://openreview.net/forum?id=rJzIBfZAb}
\BIBentrySTDinterwordspacing

\bibitem{ding2018max}
G.~W. Ding, Y.~Sharma, K.~Y.~C. Lui, and R.~Huang, ``Max-margin adversarial
  (mma) training: Direct input space margin maximization through adversarial
  training,'' \emph{arXiv preprint arXiv:1812.02637}, 2018.

\bibitem{wang2019bilateral}
J.~Wang and H.~Zhang, ``Bilateral adversarial training: Towards fast training
  of more robust models against adversarial attacks,'' in \emph{Proceedings of
  the IEEE International Conference on Computer Vision}, 2019, pp. 6629--6638.

\bibitem{zhang2019joint}
H.~Zhang and J.~Wang, ``Joint adversarial training: Incorporating both spatial
  and pixel attacks,'' \emph{arXiv preprint arXiv:1907.10737}, 2019.

\bibitem{zhang2020adversarial}
\BIBentryALTinterwordspacing
H.~Zhang and W.~Xu, ``Adversarial interpolation training: A simple approach for
  improving model robustness,'' 2020. [Online]. Available:
  \url{https://openreview.net/forum?id=Syejj0NYvr}
\BIBentrySTDinterwordspacing

\bibitem{lee2014deeplysupervised}
C.-Y. Lee, S.~Xie, P.~Gallagher, Z.~Zhang, and Z.~Tu, ``Deeply-supervised
  nets,'' 2014.

\bibitem{Zhang2019theoretically}
H.~Zhang, Y.~Yu, J.~Jiao, E.~P. Xing, L.~E. Ghaoui, and M.~I. Jordan,
  ``Theoretically principled trade-off between robustness and accuracy,''
  \emph{arXiv preprint arXiv:1901.08573}, 2019.

\bibitem{feature_scatter}
H.~Zhang and J.~Wang, ``Defense against adversarial attacks using feature
  scattering-based adversarial training,'' in \emph{Advances in Neural
  Information Processing Systems}, 2019.

\bibitem{hendrycks2019robustness}
D.~Hendrycks and T.~Dietterich, ``Benchmarking neural network robustness to
  common corruptions and perturbations,'' \emph{Proceedings of the
  International Conference on Learning Representations}, 2019.

\bibitem{goodfellow2014explaining}
\BIBentryALTinterwordspacing
I.~Goodfellow, J.~Shlens, and C.~Szegedy, ``Explaining and harnessing
  adversarial examples,'' in \emph{International Conference on Learning
  Representations}, 2015. [Online]. Available:
  \url{http://arxiv.org/abs/1412.6572}
\BIBentrySTDinterwordspacing

\bibitem{kurakin2016adversarial}
A.~Kurakin, I.~Goodfellow, and S.~Bengio, ``Adversarial machine learning at
  scale,'' \emph{arXiv preprint arXiv:1611.01236}, 2016.

\bibitem{tramer2018ensemble}
\BIBentryALTinterwordspacing
F.~Tramèr, A.~Kurakin, N.~Papernot, I.~Goodfellow, D.~Boneh, and P.~McDaniel,
  ``Ensemble adversarial training: Attacks and defenses,'' in
  \emph{International Conference on Learning Representations}, 2018. [Online].
  Available: \url{https://openreview.net/forum?id=rkZvSe-RZ}
\BIBentrySTDinterwordspacing

\bibitem{carlini2017towards}
N.~Carlini and D.~Wagner, ``Towards evaluating the robustness of neural
  networks,'' in \emph{2017 ieee symposium on security and privacy (sp)}.\hskip
  1em plus 0.5em minus 0.4em\relax IEEE, 2017, pp. 39--57.

\bibitem{shafahi2019adversarial}
A.~Shafahi, M.~Najibi, M.~A. Ghiasi, Z.~Xu, J.~Dickerson, C.~Studer, L.~S.
  Davis, G.~Taylor, and T.~Goldstein, ``Adversarial training for free!'' in
  \emph{Advances in Neural Information Processing Systems}, 2019, pp.
  3353--3364.

\bibitem{Wong2020Fast}
\BIBentryALTinterwordspacing
E.~Wong, L.~Rice, and J.~Z. Kolter, ``Fast is better than free: Revisiting
  adversarial training,'' in \emph{International Conference on Learning
  Representations}, 2020. [Online]. Available:
  \url{https://openreview.net/forum?id=BJx040EFvH}
\BIBentrySTDinterwordspacing

\bibitem{lamb2019interpolated}
A.~Lamb, V.~Verma, J.~Kannala, and Y.~Bengio, ``Interpolated adversarial
  training: Achieving robust neural networks without sacrificing accuracy,''
  \emph{arXiv preprint arXiv:1906.06784}, 2019.

\bibitem{zhang2017mixup}
H.~Zhang, M.~Cisse, Y.~N. Dauphin, and D.~Lopez-Paz, ``mixup: Beyond empirical
  risk minimization,'' \emph{arXiv preprint arXiv:1710.09412}, 2017.

\bibitem{verma2018manifold}
V.~Verma, A.~Lamb, C.~Beckham, A.~Najafi, I.~Mitliagkas, A.~Courville,
  D.~Lopez-Paz, and Y.~Bengio, ``Manifold mixup: Better representations by
  interpolating hidden states,'' \emph{arXiv preprint arXiv:1806.05236}, 2018.

\bibitem{xie2020adversarial}
C.~Xie, M.~Tan, B.~Gong, J.~Wang, A.~L. Yuille, and Q.~V. Le, ``Adversarial
  examples improve image recognition,'' in \emph{Proceedings of the IEEE/CVF
  Conference on Computer Vision and Pattern Recognition}, 2020, pp. 819--828.

\bibitem{shu2020preparing}
M.~Shu, Z.~Wu, M.~Goldblum, and T.~Goldstein, ``Preparing for the worst: Making
  networks less brittle with adversarial batch normalization,'' \emph{arXiv
  preprint arXiv:2009.08965}, 2020.

\bibitem{li2020shape}
Y.~Li, Q.~Yu, M.~Tan, J.~Mei, P.~Tang, W.~Shen, A.~Yuille, and C.~Xie,
  ``Shape-texture debiased neural network training,'' \emph{arXiv preprint
  arXiv:2010.05981}, 2020.

\bibitem{Mustafa_2019_ICCV}
A.~Mustafa, S.~Khan, M.~Hayat, R.~Goecke, J.~Shen, and L.~Shao, ``Adversarial
  defense by restricting the hidden space of deep neural networks,'' in
  \emph{The IEEE International Conference on Computer Vision (ICCV)}, October
  2019.

\bibitem{zhong2019adversarial}
Y.~Zhong and W.~Deng, ``Adversarial learning with margin-based triplet
  embedding regularization,'' in \emph{Proceedings of the IEEE International
  Conference on Computer Vision}, 2019, pp. 6549--6558.

\bibitem{li2019improving}
P.~Li, J.~Yi, B.~Zhou, and L.~Zhang, ``Improving the robustness of deep neural
  networks via adversarial training with triplet loss,'' \emph{arXiv preprint
  arXiv:1905.11713}, 2019.

\bibitem{mao2019metric}
C.~Mao, Z.~Zhong, J.~Yang, C.~Vondrick, and B.~Ray, ``Metric learning for
  adversarial robustness,'' in \emph{Advances in Neural Information Processing
  Systems}, 2019, pp. 478--489.

\bibitem{gatys2016image}
L.~A. Gatys, A.~S. Ecker, and M.~Bethge, ``Image style transfer using
  convolutional neural networks,'' in \emph{Proceedings of the IEEE conference
  on computer vision and pattern recognition}, 2016, pp. 2414--2423.

\bibitem{cohen2019certified}
J.~M. Cohen, E.~Rosenfeld, and J.~Z. Kolter, ``Certified adversarial robustness
  via randomized smoothing,'' \emph{arXiv preprint arXiv:1902.02918}, 2019.

\bibitem{Kannan2018AdversarialLP}
H.~Kannan, A.~Kurakin, and I.~J. Goodfellow, ``Adversarial logit pairing,''
  \emph{ArXiv}, vol. abs/1803.06373, 2018.

\bibitem{zantedeschi2017efficient}
V.~Zantedeschi, M.-I. Nicolae, and A.~Rawat, ``Efficient defenses against
  adversarial attacks,'' in \emph{Proceedings of the 10th ACM Workshop on
  Artificial Intelligence and Security}, 2017, pp. 39--49.

\bibitem{chen2019improving}
H.-Y. Chen, J.-H. Liang, S.-C. Chang, J.-Y. Pan, Y.-T. Chen, W.~Wei, and D.-C.
  Juan, ``Improving adversarial robustness via guided complement entropy,'' in
  \emph{Proceedings of the IEEE International Conference on Computer Vision},
  2019, pp. 4881--4889.

\bibitem{dong2018boosting}
Y.~Dong, F.~Liao, T.~Pang, H.~Su, J.~Zhu, X.~Hu, and J.~Li, ``Boosting
  adversarial attacks with momentum,'' in \emph{Proceedings of the IEEE
  conference on computer vision and pattern recognition}, 2018, pp. 9185--9193.

\bibitem{moosavi2016deepfool}
S.-M. Moosavi-Dezfooli, A.~Fawzi, and P.~Frossard, ``Deepfool: a simple and
  accurate method to fool deep neural networks,'' in \emph{Proceedings of the
  IEEE conference on computer vision and pattern recognition}, 2016, pp.
  2574--2582.

\bibitem{tashiro2020ods}
Y.~Tashiro, Y.~Song, and S.~Ermon, ``Diversity can be transferred: Output
  diversification for white- and black-box attacks,'' in \emph{Advances in
  Neural Information Processing Systems}, 2020.

\bibitem{uesato2018adversarial}
J.~Uesato, B.~O’donoghue, P.~Kohli, and A.~Oord, ``Adversarial risk and the
  dangers of evaluating against weak attacks,'' in \emph{International
  Conference on Machine Learning}.\hskip 1em plus 0.5em minus 0.4em\relax PMLR,
  2018, pp. 5025--5034.

\bibitem{croce2020reliable}
F.~Croce and M.~Hein, ``Reliable evaluation of adversarial robustness with an
  ensemble of diverse parameter-free attacks,'' in \emph{ICML}, 2020.

\bibitem{netzer2011reading}
Y.~Netzer, T.~Wang, A.~Coates, A.~Bissacco, B.~Wu, and A.~Y. Ng, ``Reading
  digits in natural images with unsupervised feature learning,'' 2011.

\bibitem{cifar}
A.~Krizhevsky, ``Learning multiple layers of features from tiny images,''
  Citeseer, Tech. Rep., 2009.

\bibitem{papernot2018cleverhans}
N.~Papernot, F.~Faghri, N.~Carlini, I.~Goodfellow, R.~Feinman, A.~Kurakin,
  C.~Xie, Y.~Sharma, T.~Brown, A.~Roy, A.~Matyasko, V.~Behzadan,
  K.~Hambardzumyan, Z.~Zhang, Y.-L. Juang, Z.~Li, R.~Sheatsley, A.~Garg,
  J.~Uesato, W.~Gierke, Y.~Dong, D.~Berthelot, P.~Hendricks, J.~Rauber, and
  R.~Long, ``Technical report on the cleverhans v2.1.0 adversarial examples
  library,'' \emph{arXiv preprint arXiv:1610.00768}, 2018.

\bibitem{croce2020minimally}
F.~Croce and M.~Hein, ``Minimally distorted adversarial examples with a fast
  adaptive boundary attack,'' in \emph{ICML}, 2020.

\bibitem{ACFH2020square}
M.~Andriushchenko, F.~Croce, N.~Flammarion, and M.~Hein, ``Square attack: a
  query-efficient black-box adversarial attack via random search,'' 2020.

\bibitem{Wu2020Defending}
\BIBentryALTinterwordspacing
T.~Wu, L.~Tong, and Y.~Vorobeychik, ``Defending against physically realizable
  attacks on image classification,'' in \emph{International Conference on
  Learning Representations}, 2020. [Online]. Available:
  \url{https://openreview.net/forum?id=H1xscnEKDr}
\BIBentrySTDinterwordspacing

\bibitem{recht2018cifar}
B.~Recht, R.~Roelofs, L.~Schmidt, and V.~Shankar, ``Do cifar-10 classifiers
  generalize to cifar-10?'' \emph{arXiv preprint arXiv:1806.00451}, 2018.

\bibitem{ILSVRC15}
O.~Russakovsky, J.~Deng, H.~Su, J.~Krause, S.~Satheesh, S.~Ma, Z.~Huang,
  A.~Karpathy, A.~Khosla, M.~Bernstein, A.~C. Berg, and L.~Fei-Fei, ``{ImageNet
  Large Scale Visual Recognition Challenge},'' \emph{International Journal of
  Computer Vision (IJCV)}, vol. 115, no.~3, pp. 211--252, 2015.

\bibitem{Darlow2018CINIC10IN}
L.~N. Darlow, E.~Crowley, A.~Antoniou, and A.~J. Storkey, ``Cinic-10 is not
  imagenet or cifar-10,'' \emph{ArXiv}, vol. abs/1810.03505, 2018.

\bibitem{motiian2017unified}
S.~Motiian, M.~Piccirilli, D.~A. Adjeroh, and G.~Doretto, ``Unified deep
  supervised domain adaptation and generalization,'' in \emph{Proceedings of
  the IEEE International Conference on Computer Vision}, 2017, pp. 5715--5725.

\bibitem{jackson2019style}
P.~T. Jackson, A.~Atapour-Abarghouei, S.~Bonner, T.~P. Breckon, and B.~Obara,
  ``Style augmentation: Data augmentation via style randomization,'' in
  \emph{Proceedings of the IEEE Conference on Computer Vision and Pattern
  Recognition Workshops}, 2019, pp. 83--92.

\bibitem{cui2020learnable}
J.~Cui, S.~Liu, L.~Wang, and J.~Jia, ``Learnable boundary guided adversarial
  training,'' \emph{arXiv preprint arXiv:2011.11164}, 2020.

\bibitem{rebuffi2021fixing}
S.-A. Rebuffi, S.~Gowal, D.~A. Calian, F.~Stimberg, O.~Wiles, and T.~Mann,
  ``Fixing data augmentation to improve adversarial robustness,'' \emph{arXiv
  preprint arXiv:2103.01946}, 2021.

\bibitem{rebuffi2021data}
S.-A. Rebuffi, S.~Gowal, D.~A. Calian, F.~Stimberg, O.~Wiles, and T.~A. Mann,
  ``Data augmentation can improve robustness,'' \emph{Advances in Neural
  Information Processing Systems}, vol.~34, pp. 29\,935--29\,948, 2021.

\bibitem{he2016deep}
K.~He, X.~Zhang, S.~Ren, and J.~Sun, ``Deep residual learning for image
  recognition,'' in \emph{Proceedings of the IEEE conference on computer vision
  and pattern recognition}, 2016, pp. 770--778.

\bibitem{Simonyan14verydeep}
K.~Simonyan and A.~Zisserman, ``Very deep convolutional networks for
  large-scale image recognition,'' 2014.

\bibitem{obfuscated-gradients}
\BIBentryALTinterwordspacing
A.~Athalye, N.~Carlini, and D.~Wagner, ``Obfuscated gradients give a false
  sense of security: Circumventing defenses to adversarial examples,'' in
  \emph{Proceedings of the 35th International Conference on Machine Learning,
  {ICML} 2018}, Jul. 2018. [Online]. Available:
  \url{https://arxiv.org/abs/1802.00420}
\BIBentrySTDinterwordspacing

\bibitem{nakano2019a}
R.~Nakano, ``A discussion of 'adversarial examples are not bugs, they are
  features': Adversarially robust neural style transfer,'' \emph{Distill},
  2019, https://distill.pub/2019/advex-bugs-discussion/response-4.

\bibitem{gowal2021improving}
S.~Gowal, S.-A. Rebuffi, O.~Wiles, F.~Stimberg, D.~A. Calian, and T.~A. Mann,
  ``Improving robustness using generated data,'' \emph{Advances in Neural
  Information Processing Systems}, vol.~34, 2021.

\bibitem{hendrycks2019using}
D.~Hendrycks, K.~Lee, and M.~Mazeika, ``Using pre-training can improve model
  robustness and uncertainty,'' in \emph{International Conference on Machine
  Learning}.\hskip 1em plus 0.5em minus 0.4em\relax PMLR, 2019, pp. 2712--2721.

\bibitem{gowal2020uncovering}
S.~Gowal, C.~Qin, J.~Uesato, T.~Mann, and P.~Kohli, ``Uncovering the limits of
  adversarial training against norm-bounded adversarial examples,'' \emph{arXiv
  preprint arXiv:2010.03593}, 2020.

\end{thebibliography}

\begin{IEEEbiography}[{\includegraphics[width=1in,height=1.5in,clip,keepaspectratio]{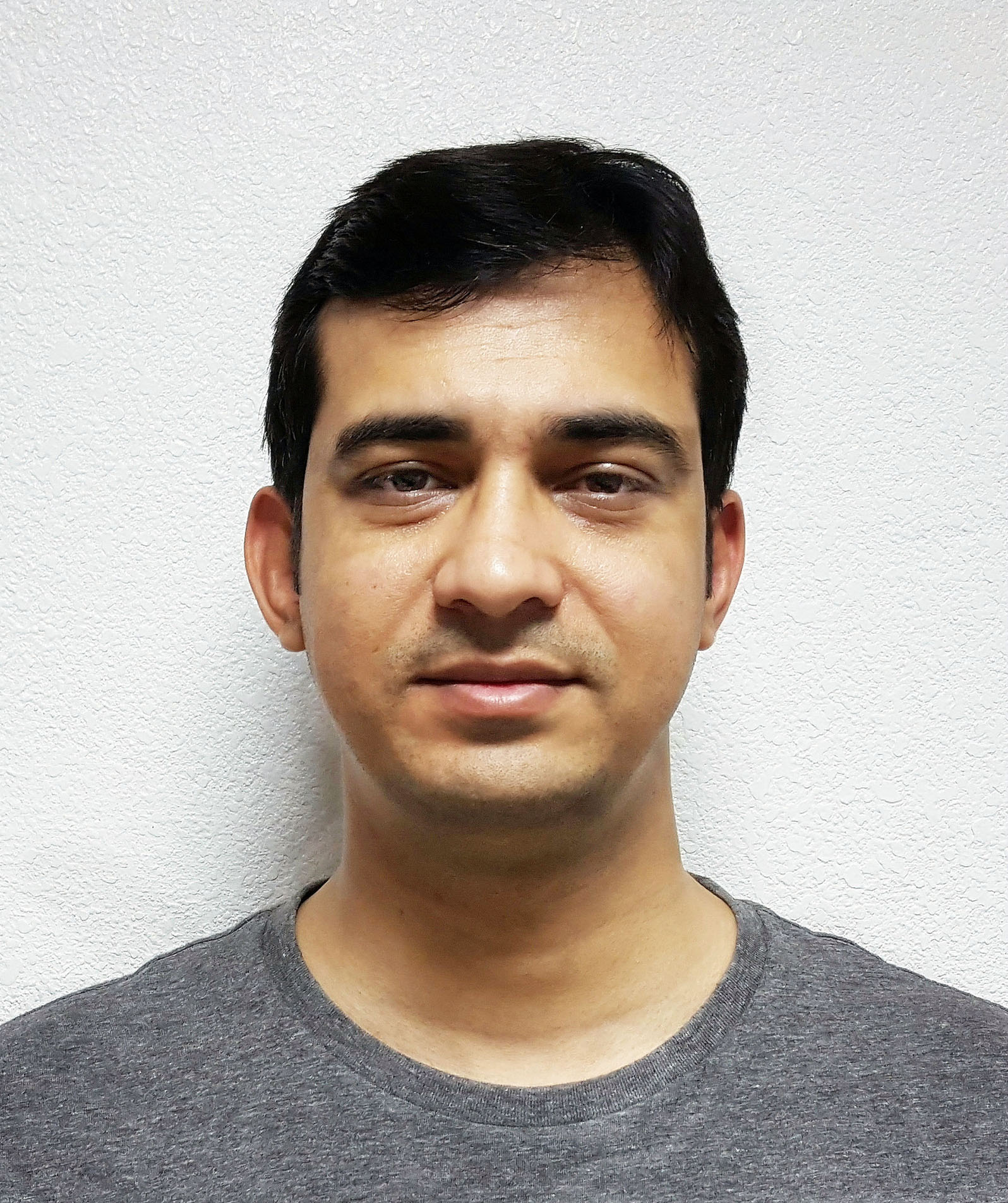}}]
{Muzammal Naseer}
received the Ph.D. degree from the Australian National University (ANU) in 2022, where he was the recipient of a competitive postgraduate scholarship. He is currently a postdoctoral researcher at Mohamed Bin Zayed University of Artificial Intelligence. He served as a researcher at Data61, CSIRO, and Inception Institute of Artificial Intelligence from 2018-2020. He has published at well recognized machine learning and computer vision venues including NeurIPS, ICLR, ICCV, and CVPR with two Oral and two spotlight presentations. He received student travel award from NeurIPS in 2019. He received Gold Medal for outstanding performance in the B.Sc. degree.
\end{IEEEbiography}

\begin{IEEEbiography}[{\includegraphics[width=1in,height=1.5in,clip,keepaspectratio]{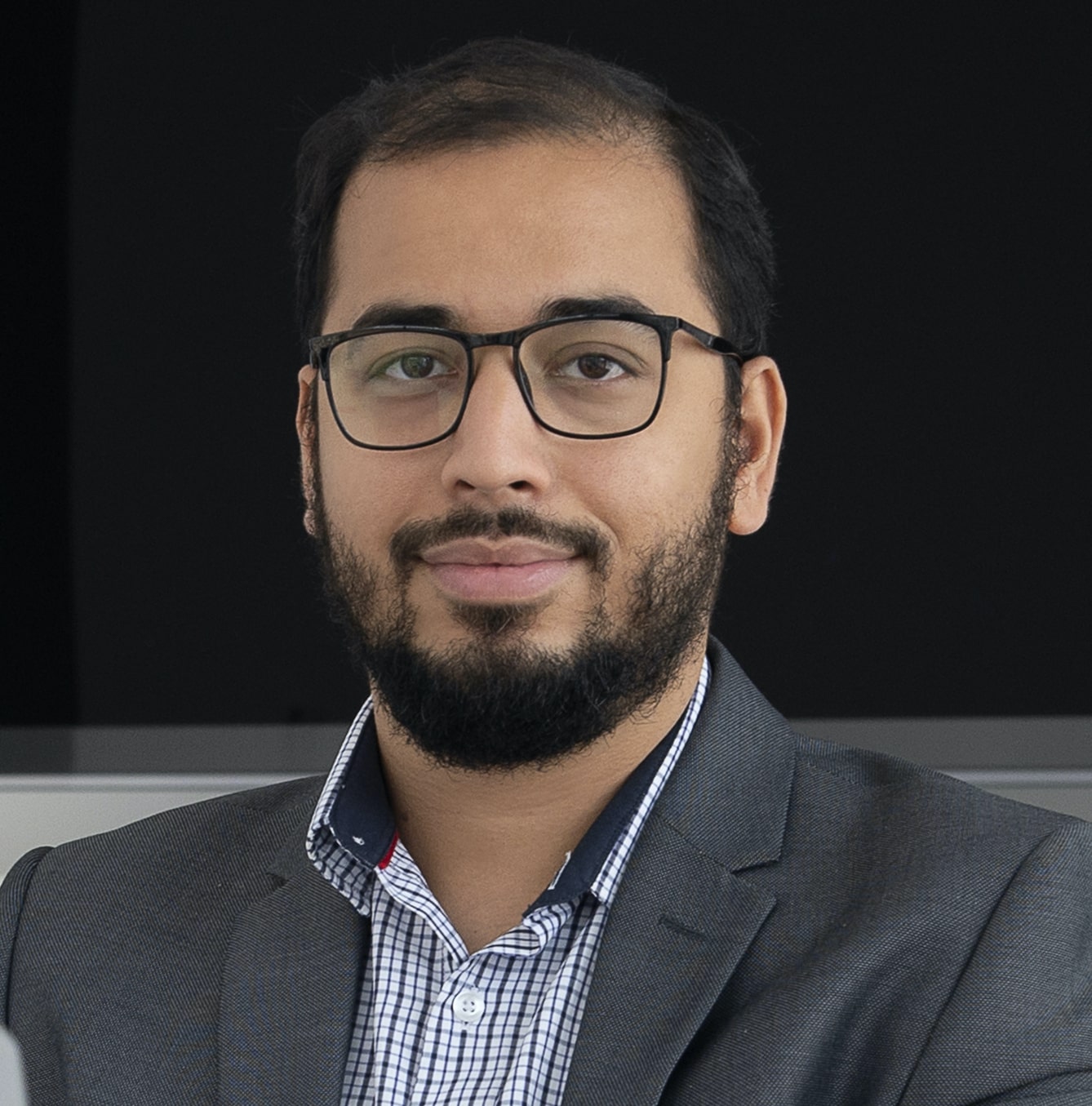}}]
{Salman Khan} (M'14-SM'22)
 received the Ph.D. degree from The University of Western Australia, in 2016. His Ph.D. thesis received an honorable mention on the Deans List Award. From 2016 to 2018, he was a Research Scientist with Data61, CSIRO. He was a Senior Scientist with Inception Institute of Artificial Intelligence from 2018-2020. He is currently acting as an Associate Professor at Mohamed Bin Zayed University of Artificial Intelligence, since 2020, and an Adjunct Lecturer with Australian National University, since 2016. He has served as a program committee member for several premier conferences, including CVPR, ICCV, IROS, ICRA, ICLR, NeurIPS and ECCV. In 2019, he was awarded the outstanding reviewer award at CVPR and the best paper award at ICPRAM 2020. His research interests include computer vision and machine learning.
 \end{IEEEbiography}

\begin{IEEEbiography}[{\includegraphics[width=1in,height=1.5in,clip,keepaspectratio]{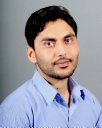}}]
{Munawar Hayat}
 received his PhD from The University of Western Australia (UWA). His PhD thesis received multiple awards, including the Deans List Honorable Mention Award and the Robert Street Prize. After his PhD, he joined IBM Research as a postdoc and then moved to the University of Canberra as an Assistant Professor. He is currently a Senior Lecturer at Monash University, Australia. Munawar was granted two US patents, and has published over 30 papers at leading venues in his field, including TPAMI, IJCV, CVPR, ECCV and ICCV. His research interests are in computer vision and machine/deep learning.
 \end{IEEEbiography}

\begin{IEEEbiography}[{\includegraphics[width=1in,height=1.5in,clip,keepaspectratio]{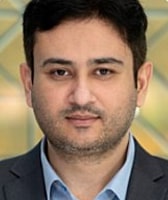}}]
{Fahad Shahbaz Khan}
is a faculty member at MBZUAI, United Arab Emirates and Linköping University, Sweden. He received the M.Sc. degree in Intelligent Systems Design from Chalmers University of Technology, Sweden and a Ph.D. degree in Computer Vision from Autonomous University of Barcelona, Spain. He has achieved top ranks on various international challenges (Visual Object Tracking VOT: 1st 2014 and 2018, 2nd 2015, 1st 2016; VOT-TIR: 1st 2015 and 2016; OpenCV Tracking: 1st 2015; 1st PASCAL VOC 2010). He received the best paper award in the computer vision track at IEEE ICPR 2016. His research interests include a wide range of topics within computer vision and machine learning, such as object recognition, object detection, action recognition and visual tracking. He has published over 100 conference papers, journal articles, and book contributions in these areas. He has served as a guest editor of IEEE Transactions on Pattern Analysis and Machine Intelligence, IEEE Transactions on Neural Networks and Learning Systems and is an associate editor of Image and Vision Computing Journal. He serves as a regular program committee member for leading computer vision conferences such as CVPR, ICCV, and ECCV.
\end{IEEEbiography}

\begin{IEEEbiography}[{\includegraphics[width=1in,height=1.5in,clip,keepaspectratio]{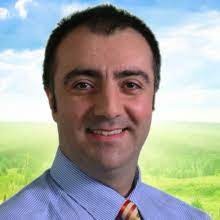}}]
{Fatih Porikli}
(M’96–SM’04–F’14) received the Ph.D. degree from New York University, in 2002. He was a Distinguished Research Scientist with Mitsubishi Electric Research Laboratories. He was a full tenured Professor in the Research School of Engineering, Australian National University, and a Chief Scientist with the Global Media Technologies Lab, Huawei, Santa Clara. He is currently the Global Lead of Perception at Qualcomm. He has authored over 300 publications, co-edited two books, and invented 66 patents. His research interests include computer vision, pattern recognition, manifold learning, image enhancement, robust and sparse optimization, and online learning with commercial applications in video surveillance, car navigation, robotics, satellite, and medical systems. He was a recipient of the Research and Development 100 Scientist of the Year Award, in 2006. He received five best paper awards at premier IEEE conferences and five other professional prizes. He is serving as an associate editor for several journals for the past 12 years. He has also served in the organizing committees of several flagship conferences, including ICCV, ECCV, and CVPR.
\end{IEEEbiography}






\end{document}